\documentclass{elsarticle} 

\usepackage{lineno}

\usepackage{times}
\usepackage{helvet}
\usepackage{courier}
\usepackage{amssymb}
\usepackage{latexsym}
\usepackage{amsmath}
\usepackage{multirow}
\usepackage{url}
\usepackage{amsmath}
\usepackage{epsf}
\usepackage{epsfig}
\usepackage{graphicx}
\usepackage{color}
\usepackage{wrapfig}
\usepackage{colortbl}
\usepackage{verbatim}
\usepackage{subfigure}
\usepackage{multirow,rotating}
\usepackage{xcolor} 
\usepackage{pdfpages}

\usepackage{algorithm}
\usepackage{algpseudocode}
\usepackage{amssymb}
\usepackage{amsmath}
\usepackage{lscape}

\modulolinenumbers[5]

\journal{Neurocomputing}

\begin{document}

\begin{frontmatter}

\title{Deep Supervised/Reinforcement Learning for Interactive Robots Playing Social Games}
\title{Multimodal Deep Supervised/Reinforcement Learning for Interactive Robots}
\title{A Deep Learning Approach for Multimodal Social Robots}
\title{A \textcolor{black}{Data-Efficient} Deep Learning Approach for \textcolor{black}{Deployable} Multimodal Social Robots}

\author{Heriberto Cuay\'ahuitl}
\address{University of Lincoln, School of Computer Science,\\
Lincoln Centre for Autonomous Systems (L-CAS)}



\address[mymainaddress]{Brayford Pool, Lincoln, LN6 7TS, United Kingdom}

\begin{abstract}
\textcolor{black}{The deep supervised and reinforcement learning paradigms (among others) have the potential to endow interactive multimodal social robots with the ability of acquiring skills autonomously. But it is still not very clear yet how they can be best deployed in real world applications. As a step in this direction, we propose a deep learning-based approach for efficiently training a humanoid robot to play multimodal games---and use the game of `Noughts \& Crosses' with two variants as a case study. Its minimum requirements for learning to perceive and interact are based on a few hundred example images, a few  example multimodal dialogues and physical demonstrations of robot manipulation, and automatic simulations. In addition, we propose novel algorithms for robust visual game tracking and for competitive policy learning with high winning rates, which substantially outperform DQN-based baselines. While an automatic evaluation shows evidence that the proposed approach can be easily extended to new games with competitive robot behaviours, a human evaluation with 130 humans playing with the {\it Pepper} robot confirms that highly accurate visual perception is required for successful game play.\footnote{\url{https://doi.org/10.1016/j.neucom.2018.09.104}}}
\end{abstract}


\end{frontmatter}


\section{Introduction}
\textcolor{black}{In a not so distant future, we will be able to buy purposeful and socially-aware humanoid robots that can be delivered home much like buying personal computers nowadays. While robots may come with a pre-defined or pre-trained set of skills---arguably and ideally---they should be able to self-adapt or self-extend for carrying out new useful tasks relevant to their individual user(s). A new task can be one that is entirely distinct from pre-defined skills or one that is similar but not the same. We will refer to both types as `new tasks' and present two examples of new tasks in our case study below. Deploying robots with pre-built skills is still challenging assuming that they have to adapt to different spatial and social environments each time an adaptation of existing knowledge occurs. Deploying robots with the ability to acquire new skills has the potential to be even more challenging. This latter form of deployment is of great interest to AI because it requires advanced multimodal communication (via human-like verbal and non-verbal commands) in order to achieve a task or set of tasks successfully. A concrete scenario e.g. is as follows: Your new robot has arrived and you want to teach it to play a game (that is at least partially unknown to the robot) so it can play with you, your family and friends whenever you want. Having said that ... {\it How can the robot be equipped and/or trained to play such a game with a reduced amount of human intervention?} {\it What are the basic building blocks required to make that happen?} This article makes a step towards answering some of these challenging questions and discusses future work towards purposeful and socially-aware humanoid robots.}


\begin{figure}[t!]
     \begin{center}
             \subfigure[Nougths \& Crosses]{%
            \includegraphics[angle=90,width=0.4435\textwidth]{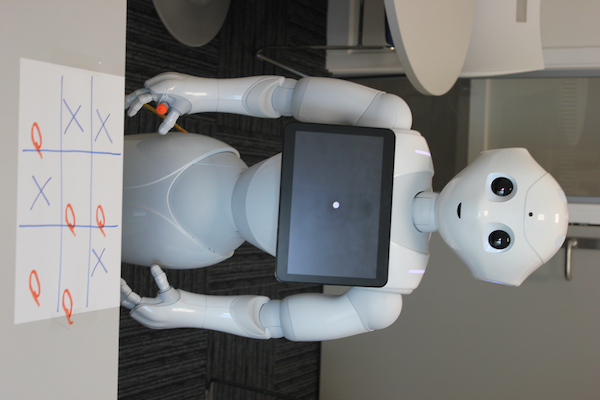}
        }
        \subfigure[Ultimate Nougths \& Crosses]{%
            \includegraphics[width=0.374\textwidth]{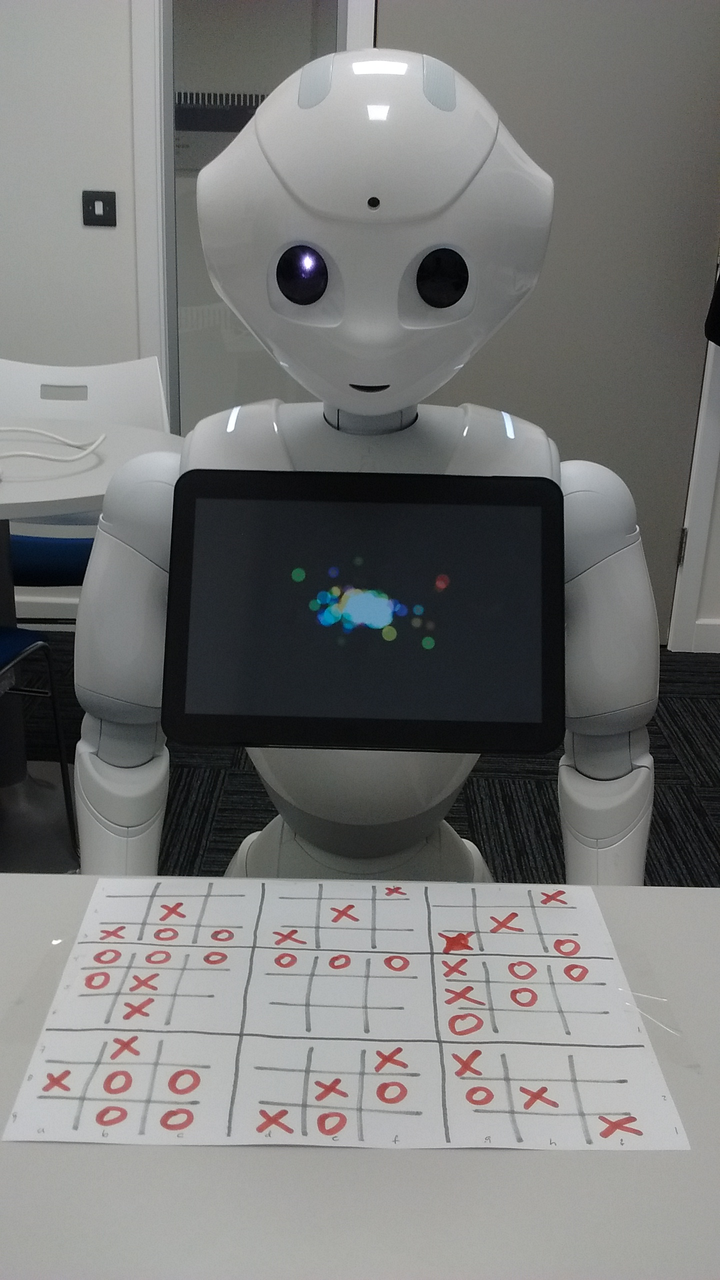}
        }
    \end{center}
   \caption{A humanoid robot playing the game of noughts and crosses with two variants using multiple modalities and learnt behaviour}
   \label{MultimodalGames}
\end{figure}

\begin{figure}[t]	
     \begin{center}
             \subfigure[Robot game move1]{%
            \includegraphics[width=0.22\textwidth]{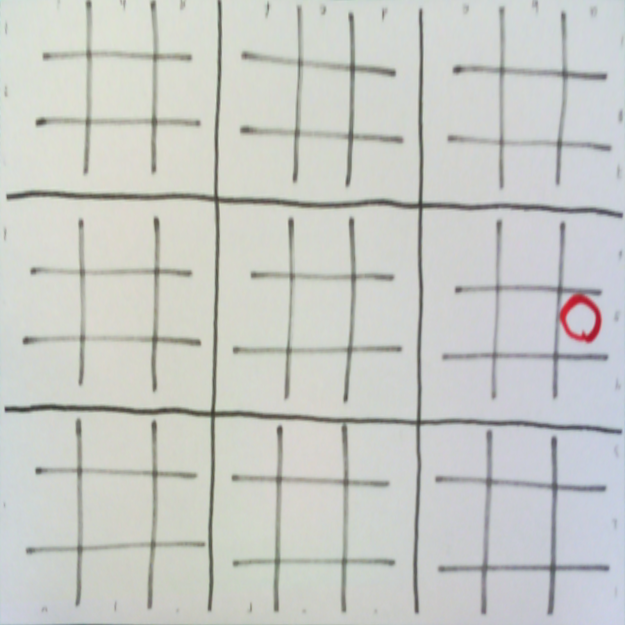}
        }
             \subfigure[Human game move1]{%
            \includegraphics[width=0.22\textwidth]{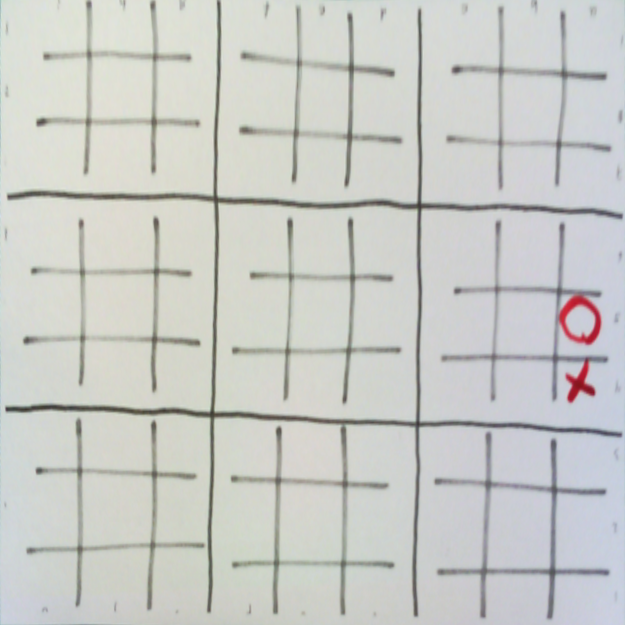}
        }
        \subfigure[Robot game move2]{%
            \includegraphics[width=0.22\textwidth]{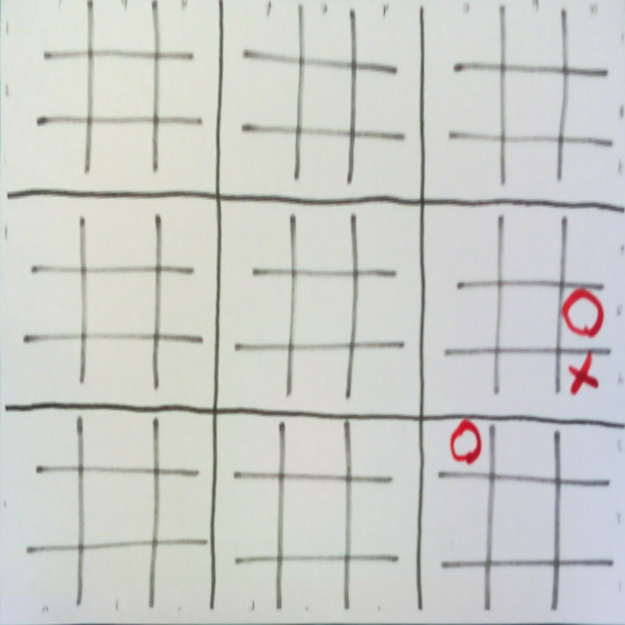}
        }
        \subfigure[Human game move2]{%
            \includegraphics[width=0.22\textwidth]{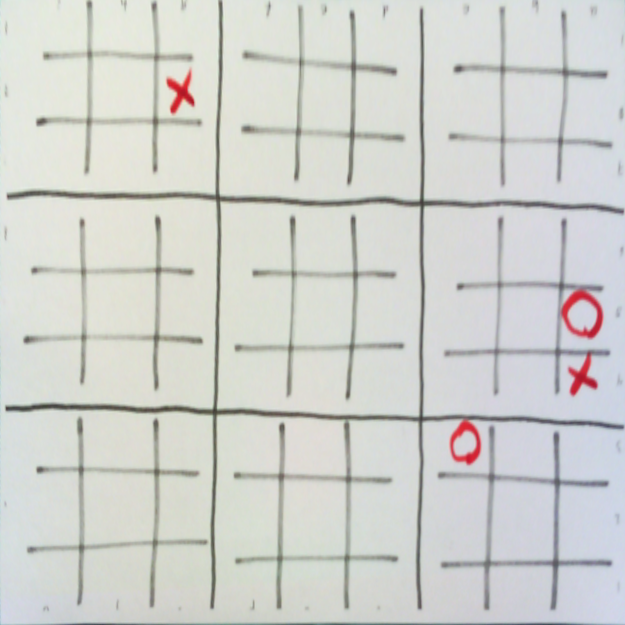}
        }
        \dots
    \end{center}
\caption{\label{GameGrid}Example robot and user game moves --- robot's field of view from bottom to top}
   \label{GameMoves}
\end{figure}

As a case study, the multimodal game that we focus on is {\it Noughts and Crosses} also known as `Tic-Tac-Toe'---with two variants. In both games players alternate turns, and each player is represented by either noughts or crosses.
\begin{itemize}
\item Its standard version uses a 3$\times$3 grid, where a game is won if and only if three noughts or crosses are in line or diagonal (a draw otherwise)---see Figure~\ref{MultimodalGames} (left). One player adopts noughts and the other crosses, alternating turns until the game is over. In this game, the more expertise the players acquire, the more likely it is to end up in a draw. This has motivated the development of variants of the game with higher degrees of complexity. 
\item A substantially more difficult variant, called {\it Ultimate Noughts and Crosses}\footnote{\url{https://en.wikipedia.org/wiki/Ultimate_tic-tac-toe}}, uses 3$\times$3 subgrids each of 3$\times$3 squares (81 squares in total), where the goal is to win three subgrids (each of 3$\times$3 squares) in line or diagonal. In this latter game, while the first game move is allowed to take any of the 81 squares, a subsequent game move is restricted to take a square in the subgrid that mirrors the previous game move of the opponent player. For example, a player taking the middle right square of any subgrid would restrict the opponent to take a move anywhere in the middle right subgrid---as shown in Figure~\ref{GameMoves}(a) and (b). Similarly, a player taking for example the bottom-right square of any subgrid would restrict the opponent to take a move anywhere in the bottom-right subgrid---as in Figure~\ref{GameMoves}(b) and (c), and so on. There is one exception to these restrictions: a player can take an empty square anywhere in the entire board if and only if the target subgrid has been won/lost/draw already. This game is more advanced than its standard counterpart and strategically challenging---see Figure~\ref{MultimodalGames} (right). 
\end{itemize}

The challenging task for the robot is to successfully play either game against unknown humans and partially familiar physical environments. \textcolor{black}{This article describes a machine intelligence approach for efficiently training such a deployable robot, and makes the following contributions:
\begin{enumerate}
\item We propose a deep learning-based approach for training a multimodal robot with low data requirements. This is demonstrated by a scenario with two variants of different complexity and the following data requirements: a few hundred example images, a dozen example multimodal dialogues (see example in \cite{Cuayahuitl2017humanoids}-Appendix A), a few example physical demonstrations of handwriting, and automatically generated simulated games. Applying our approach to other games or tasks would require similar resources (though with further training examples depending on task complexity), plus a mechanism to let the robot know about valid actions and when a task has been achieved or not (e.g. game won/lost). The latter together with more refined manipulation or locomotion would require additional programming, which future work should try to automate.
\item We propose two novel learning algorithms, one for visual perception and the other one for policy learning. The former is useful for tracking the game state, i.e. what moves have been made so far by each player. Accurate recognition is important for playing the game so that the robot's view of the world is as accurate as possible, as opposed to a blurry view that would lead to unexpected or non-human-like behaviours. Deep learning can be used to provide the robot with game moves (in our case), and it can also provide the interaction agent with learnt internal representations. The latter algorithm for interaction (policy learning) is important for training an autonomous robot that learns---in a scalable way---its competitive behaviour from trial and error. Our newly proposed algorithms win substantially more than the well-known DQN method \cite{mnih-dqn-2015}. 
\item We carried out a near real world evaluation of our deep learning-based humanoid robot, who played the game of Noughts and Crosses against 130 different individuals in the wild. While most previous work has carried out evaluations using simulations only or controlled experiments in lab environments, we believe it is important and timely to show that newly developed approaches or algorithms work (to a large extent) in real or near-real world settings -- out of lab conditions. Robots interacting in the wild have to be able to deal with unstructured interactions, partially known environments, unseen human behaviour, etc. This is a big challenge for multimodal robots, and this work reports a step in this direction.
\end{enumerate}
}

\section{Related Work}

So far the topic of deep learning-based conversational and/or multimodal social robots is in many respects unexplored. Some exceptions include the following. \cite{NodaASO14} train a humanoid robot to carry out the following object manipulation behaviours: ball lift, ball roll, bell right (left and right), ball roll on a plate, and ropeway. To train this multimodal robot three neural networks are used: first, a deep autoencoder is used for feature learning from audio signals in the form of spectrograms; second, a deep autoencoder is used for feature learning from 2D images; and third, a deep autoencoder is also used for multimodal feature learning from audio and visual features generated by the previous two autoencoders. The latter learnt features are given as input to a multiclass Support Vector Machine classifier in order to predict the object manipulation task to carry out. Focusing more on social skills, \cite{QureshiNYI16} train a humanoid robot with social skills whose goal is to choose one of four actions: wait, look towards human, wave hand, and handshake. The authors use the DQN method \cite{mnih-dqn-2015} and a two-stage approach. While the first stage collects grayscale and depth images from the environment, the second stage trains two Convolutional neural nets with fused features. The robot receives a reward of +1 for a successful handshake, -0.1 for an unsuccessful handshake, and 0 otherwise. Combining social and action learning, \cite{CuayahuitlCO16,Cuayahuitl2017humanoids} train a robot to play games also using the DQN method and a variant of it. In this work a Convolutional neural net is used to predict game moves, and a fully-connected neural net is used to learn multimodal actions (18 in total) based on game rewards. Other previous works have addressed multimodal deep learning but in non-conversational settings \cite{WermterWEPEP04,NgiamKKNLN11,SrivastavaS14,LevineFDA15}. From all these works it can be observed that learning agents use small sets of actions in single-task scenarios. Thus, humanoid social robots with more complex behaviours including larger sets of actions remain to be investigated.

There is a similarly limited amount of previous work on humanoid robots playing games against human opponents. Notable exceptions include \cite{Bentivegna2004}, where the DB humanoid robot learns to play air hockey using a Nearest Neighbour classifier; \cite{short:cheating}, where the Nico humanoid torso robot plays the game of rock-paper-scissors using a `Wizzard of Oz' setting; \cite{KoberGM2012}, where the Sky humanoid robot plays catch and juggling using inverse kinematics and induced parameters with least squares linear regression; \cite{Belpaeme2013}, where the Nao robot plays a quiz game, an arm imitation game, and a dance game using tabular reinforcement learning; \cite{KimS14}, where the Genie humanoid robot plays the poker game using a `Wizard of Oz' setting; and \cite{Barakova2018}, where the NAO robot plays Checkers using a MinMax search tree. Most of these robots only exhibit non-verbal abilities and are either teleoperated or based on heuristic methods, which suggests that verbal abilities in autonomous trainable robots playing games are underdeveloped. Apart from \cite{CuayahuitlCO16,Cuayahuitl2017humanoids}, we are not aware of any other previous work in humanoid robots playing social games against human opponents and trained with deep learning methods. 

\textcolor{black}{Previous work on multimodal robots trained to carry out specific tasks and that have been deployed in the wild are almost absent as pointed out by \cite{Jung2018,HC2015aisb,Cuayahuitl2015csl}---perhaps due to the complexity involved. Most previous multimodal trainable robots are either trained and tested in simulation, or trained (usually offline) and tested in controlled conditions and/or using recruited participants. For the sake of clarity, we refer to robots deployed in the wild as those robots interacting with non-recruited participants in a non-controlled manner and with rather spontaneous, unrestricted and untimed interactions. The closest previous work is the Minerva robot \cite{Thrun_2000}, which gave 620 tours to people through the exhibitions of a museum. Another related work is the Nao robot \cite{Bohus_2014}, which gave route instructions to employees and visitors of a company building. Lessons learnt by these works include the application of probabilistic approaches to deal with uncertainty in the interaction, and challenges in starting and finishing conversational engagements with out-of-domain responses. Our work complements previous work by showcasing a robot that carries out a joint activity with people, namely playing multimodal games, in a spontaneous and uncontrolled setting.}

In the remainder of the article we describe a deep learning-based approach for efficiently training a robot with the ability of behaving with reasonable performance in a near real world deployment. In particular, we measure the effectiveness of neural-based game move interpretation and the effectiveness of Deep Q-Networks (DQN) \cite{mnih-dqn-2015} for interactive social robots. Field trial results show that the proposed approach can induce reasonable and competitive behaviours, especially when they are not affected by unseen noisy conditions.

\section{Proposed Learning Approach} 
\textcolor{black}{Our proposed approach uses two deep learning tasks in cascade with low data requirements, which is useful to enable robots with new skills and where training data is either absent or scarce.} While the first learning task predicts what is going on in the environment---game moves and who said what, the second learning task inherits such predictions in order to decide what to do or say next. Our approach is motivated by the fact that once a robot system is trained, it is expected to operate not only in known environments but also in partially-known environments. The latter may include unseen rooms, unseen furniture, and unseen human opponents, among others.
\textcolor{black}{This approach can be applied to other tasks beyond the case study in this article through the following methodology:\\
\fcolorbox{black}[HTML]{E9F0E9}{\parbox{\textwidth}{%
\begin{enumerate}
\item Collect a modest set of example images (e.g. a few hundred or as needed) and label them.
\item Train a deep supervised learner for visual perception to keep track of the environment dynamics as described in Section~\ref{Learning2Perceive}.
\item Write a set of example dialogues (e.g. a dozen or as needed) as in the Appendix of  \cite{Cuayahuitl2017humanoids}, which can be used for generating simulated interactions as in \cite{HC2016iwsds}.
\item Use the outputs of the previous two steps for training a deep reinforcement learner using simulations as described in Section~\ref{Learning2Interact}.
\item Collect a set of pre-recorded motor motions or train a component to carry out commands such as `write cross or circle in a particular location'. 
\item Test the robot using the previous resources, and iterate from step 1 if needed.
\item Deploy the robot subject to successful interactions in the previous step.
\end{enumerate}}}
}

\begin{figure}[t!]
     \begin{center}
            \includegraphics[width=0.95\textwidth]{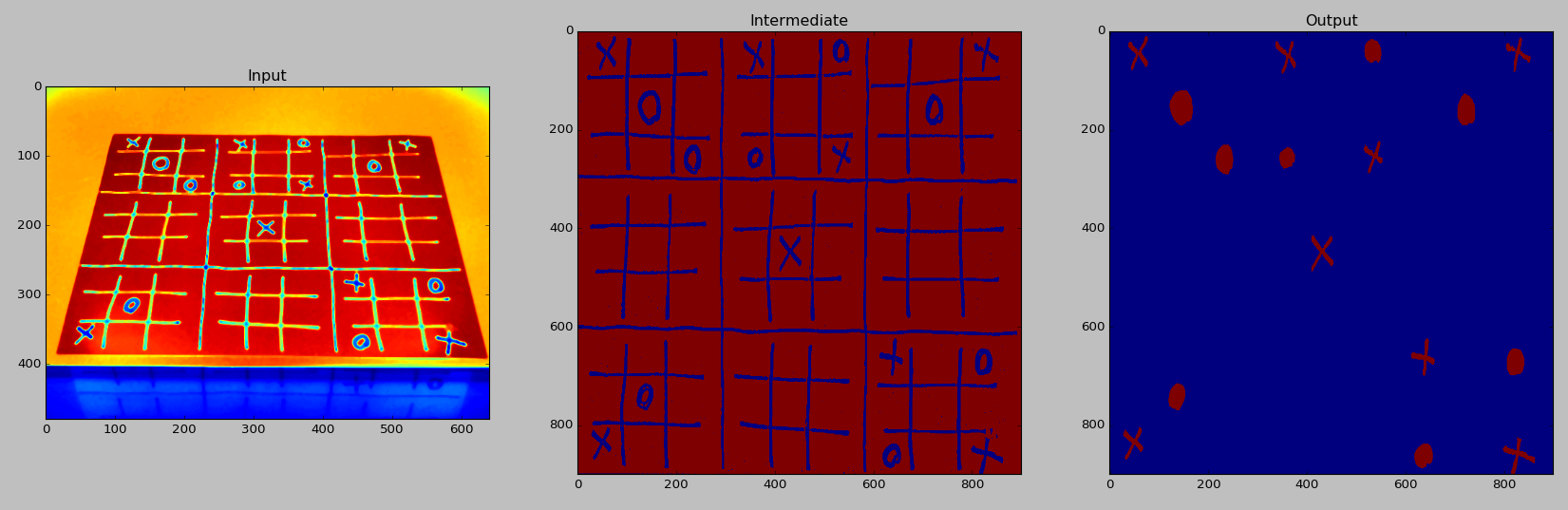}
    \end{center}
\caption{\label{GameGrid1}Raw input image in colour space BGR-Gray (left), largest contour with projected transformation (middle), and game space used for game move recognition (right). Our image pre-processing is based on OpenCV (\url{http://opencv.org/})}
   \label{images}
\end{figure}



\subsection{Visual-Based Perception of Game Moves} 
\label{Learning2Perceive}
We use the 2D camera in the robot's mouth to recognise drawings on the game grid using Algorithm~\ref{GMR}. 
The robot extracts video frames, locates the game grid using colour-based detection of the largest contour---see Figure~\ref{GameGrid1}, and splits the game grid into a set of subimages (one for each grid square) in order to pass them to a probabilistic classifier for game move recognition.

\begin{algorithm} [t!]
\caption{\label{GMR} Game Move Recogniser} 
\begin{algorithmic}[1]
\State Input: video stream from 2D camera, $n \times n$=grid size (e.g. 3x3, 9x9), noise threshold $\tau$, $\rho$=resolution of subimages, $\gamma$=pause between recognition events, $\mathcal{C}$=statistical classifier, initialise $L^{(i)}_t$=labels at time $t$ for grid square $i$
\State Output: set $\mathcal{G}$ of recognized game moves
\Repeat
   \State $F \leftarrow$ extract video frame from video stream
   \State $P \leftarrow$ extract page region from $F$
   \State $P' \leftarrow$ get projected transformation from $P$
   \If {$dist$($P'_t$,$P'_{t-1}$)+$dist$($P'_{t-1}$, $P'_{t-2}$) $>$ $\tau$} 
	 \State {\bf continue} (detect and omit hands in handwriting)
   \EndIf
   \State $G \leftarrow$ remove grid and convert image $P'$ to grayscale
   \State $G' \leftarrow$ divide $G$ into $n \times n$ images of $\rho$ pixels
   \State $L^{(i)}_{t-2} \leftarrow L^{(i)}_{t-1}$ 
   \State $L^{(i)}_{t-1} \leftarrow L^{(i)}_{t}$ 
   \State $L^{(i)}_t \leftarrow$ predict labels for each image in $G'$ using $\mathcal{C}$
   \State $i^*$ = $\arg\max_{i} \frac{1}{Z}$ $Pr(L^{(i)}_{t-2})$+$Pr(L^{(i)}_{t-1})$+$Pr(L^{(i)}_{t})$
   \If {label of grid square $i^*$ is not `nothing'} 
     \State Update $\mathcal{G}$ with game move of grid square $i^*$
	 \State {\bf break}
   \EndIf
   \State sleep $\gamma$ milliseconds
\Until end of game turn
\end{algorithmic}
\end{algorithm}

Although a game move recogniser can be trained in an online fashion, our algorithm below assumes the existence of a supervised learner trained offline. Thus, we leave the topic of online training, during the course of the interaction, as future work. In our case we use a Convolutional Neural Network (CNN) \cite{lecun-gradientbased-learning-applied-1998} $\mathcal{C}$ trained from data set $\mathcal{D}=\{({\bf x}_1,y_1),...,({\bf x}_N,y_N)\}$, where ${\bf x}_i$ are $n \times n$ matrices of pixels and $y_j$ are class labels. This classifier maps images to labels---in our case \{`nought', `cross', `nothing'\}. 
Our CNN uses the following architecture: input layer of 40$\times$40 pixels, convolutional layer with 8 filters, ReLU, pooling layer of size 2$\times$2 with stride 2, convolutional layer with 16 filters, ReLU, pooling layer of size 3$\times$3 with stride 3, and the output layer used a linear Support Vector Machine with 3 labels. This CNN is used multiple times---once per grid cell---to predict the state of each game grid, in each game turn, for detecting drawing events. Note that the larger the grid the larger the number of recognition events needed for predicting the state of the entire game grid. In addition, rather than using only the most recent video frame for game move recogition, we use a history of user and robot game moves (denoted as $\mathcal{G}$) to focus recognition on valid game moves, i.e. newly recognised moves from empty grid squares to non-empty grid squares. This process needs to be done in (near) real-time for exhibiting smooth human-robot interactions. 

\begin{figure}[t!]
     \begin{center}
             \subfigure[Before game move]{%
            \includegraphics[width=0.25\textwidth]{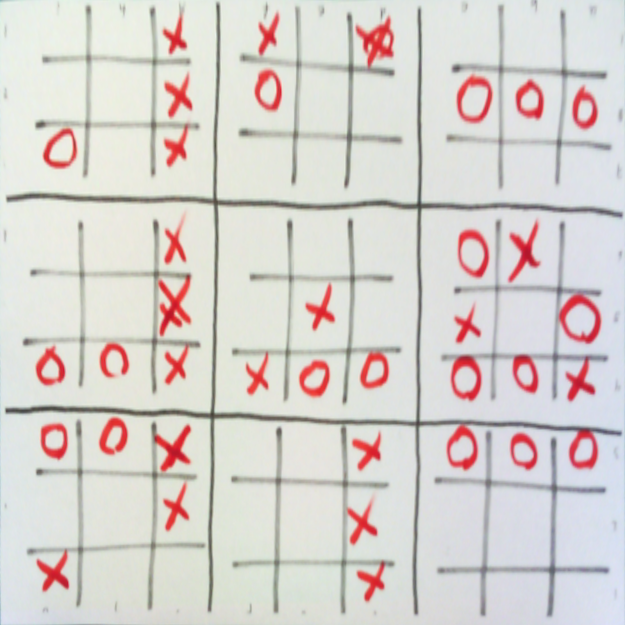}
        }
             \subfigure[During game move]{%
            \includegraphics[width=0.25\textwidth]{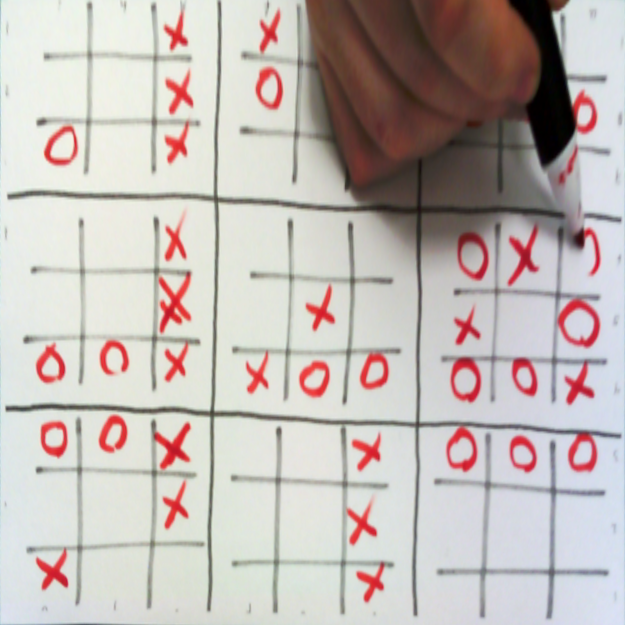}
        }
        \subfigure[After game move]{%
            \includegraphics[width=0.25\textwidth]{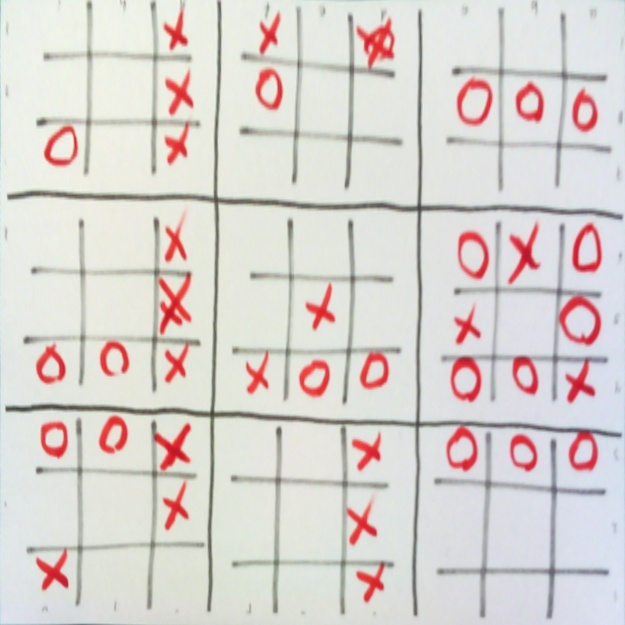}
        }
    \end{center}
\caption{\label{GameGrid}Illustration of a game move before and after handwriting}
   \label{hand}
\end{figure}

Algorithm~\ref{GMR} formalises the description above for game move recognition, which can be used at each game turn in a game. Lines 7-9\footnote{Function $dist(\cdot,\cdot)$ in Algorithm~\ref{GMR} is based on the Euclidean distance.} are particularly useful for ignoring video frames with human hands, which can be a source of misrecognitions---see Figure~\ref{hand}. In this way the game move recogniser is responsible for maintaining, as accurately as possible, the state of the game based on a set of user and robot game moves\footnote{While 18 game moves, i.e. 9 grid squares $\times$ 2 players, are used to describe the state of the standard noughts and crosses game grid, $81\times2$=162 game moves are used to describe the state of the ultimate noughts and crosses game grid. Considering a window history of 3 time steps, $18\times3$=54 and $162\times3$=486 classification events are needed at each time step for both games, respectively. In our case, the number of window histories is indefinite because human players can take a turn in their own time.} in the following format: $[who$=$usr \wedge what$=$draw \wedge where$=$i^*]$.



\subsection{Learning to Interact given Multimodal Inputs}
\label{Learning2Interact}
The visual perceptions above plus words raised in the interaction by both conversants\footnote{This work assumes that words raised by the human opponent are derived from the top recognition hypothesis of a speech recogniser. An alternative representation would be the use of word embeddings to deal with unseen words and similar meanings \cite{CuayahuitlEtAl2018slt}.} are given as input to a reinforcement learning agent to induce its behaviour, where such multimodal perceptions are mapped to multimodal actions by maximizing a long-term reward signal. 
The goal of a reinforcement learner is to find an optimal policy denoted as $\pi^{*}(s) = \arg \max_{a \in A} Q^*(s,a)$,  
%
where $Q^*$ represents the maximum sum of rewards $r_t$ discounted by factor $\gamma$ at each time step, after taking action $a$ in state $s$ \cite{SuttonB2018,Szepesvari:2010}. 
While reinforcement learners take stochastic actions during training, they select the best actions $a^*=\pi^*(s)$ at test time.

Our reinforcement learning agents approximate $Q^*$ using a multilayer neural network as in \cite{mnih-dqn-2015}. The $Q$ function is parameterised as $Q(s,a;\theta_i)$, where $\theta_i$ are the parameters (weights) of the neural network at iteration $i$. Furthermore, training a deep reinforcement learner requires a dataset of experiences $D=\{e_1,...e_N\}$ (also referred to as `experience replay memory'), where every experience is described as a tuple  $e_t=(s_t,a_t,r_t,s_{t+1})$. Inducing $Q^*_\theta$ consists in iteratively applying Q-learning updates over minibatches of experience $MB=\{(s,a,r,s')\sim U(D)\}$ drawn uniformly at random from the full data set $D$. A learning update at iteration $i$ is thus defined according to the following loss function 
\begin{equation}\nonumber
L_i(\theta_i)=\mathbb{E}_{MB} \left[ (r+\gamma \max_{a'} Q(s',a';\overline{\theta}_i)-Q(s,a;\theta_i))^2 \right],
\end{equation}  
where $\theta_i$ are the parameters of the network at iteration $i$, and $\overline{\theta}_i$ are the target parameters of the network at iteration $i$. 
The latter are held fixed between individual updates. 
This process is known as {\it Deep Q-Learning with Experience Replay} \cite{mnih-dqn-2015}. 

\begin{figure}[t!]
  \begin{center}
    \includegraphics[width=0.98\textwidth]{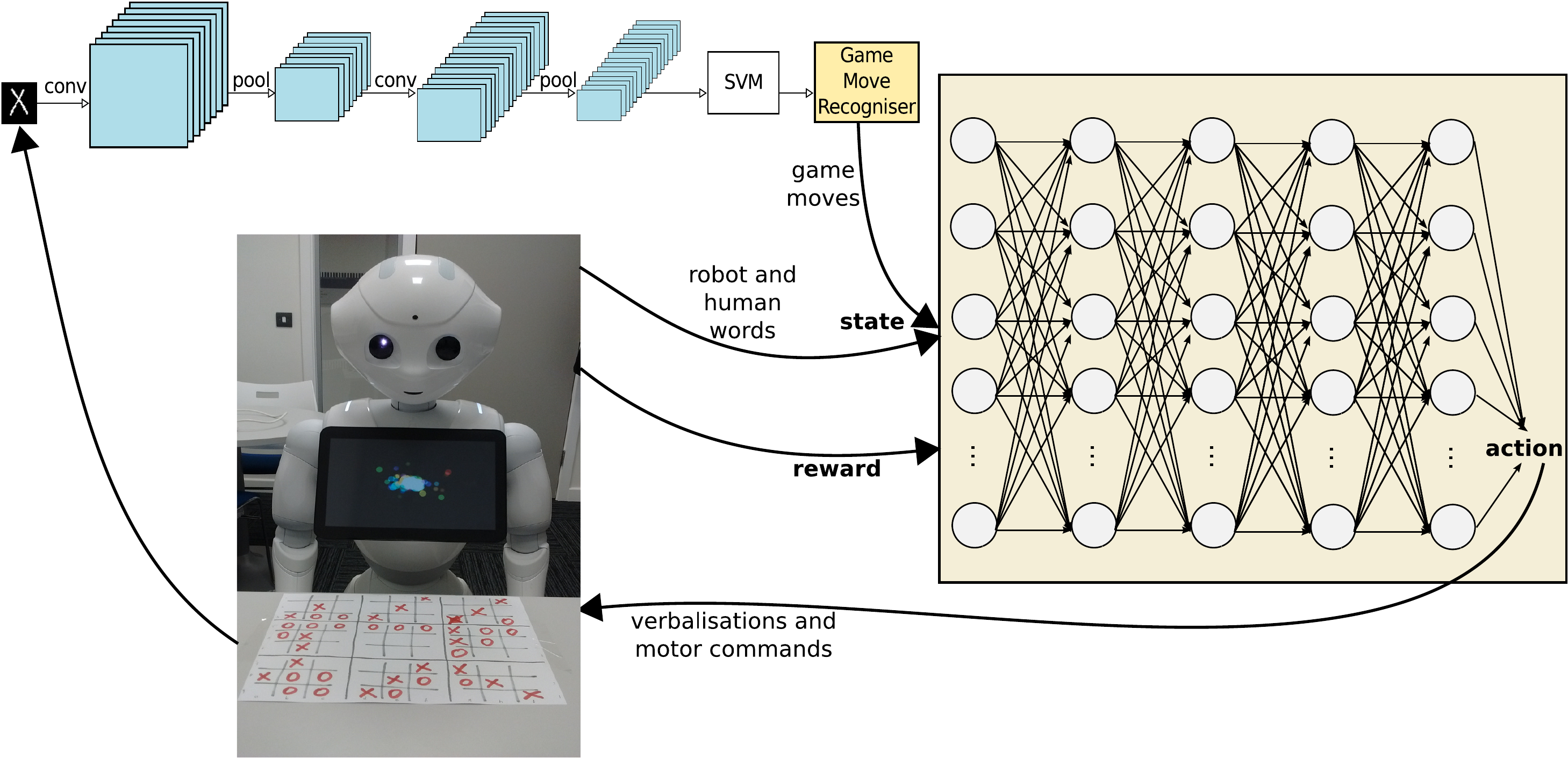}
\caption{\label{integratedSystem} Illustration of our Multimodal Deep Reinforcement Learning agent}
  \end{center}
\end{figure}

\begin{algorithm} [t!]
\caption{\label{ourAlgorithm} Competitive DQN Learning}\label{ndqn} 
\begin{algorithmic}[1]
\State Initialise Deep Q-Networks with replay memory $D$, action-value function $Q$ with random weights $\theta$, and target action-value functions $\hat{Q}$ with weights $\hat{\theta}=\theta$
\Repeat
   \State $s \leftarrow$ initial environment state in $S$
   \Repeat 
      \State $\hat{A}= 
\begin{cases}
    \text{actions with } \min(r(s,a)<0\ \forall a \in A)\\
     \emptyset \text{  otherwise}
\end{cases}$
      \State $A= 
\begin{cases}
    \text{action(s) leading to win (legally) the game}\\
     \text{all available actions in state $s$} \text{  otherwise}
\end{cases}$
      \State $a= 
\begin{cases}
    rand_{a \in A} \text{ if } \mbox{random number} \le \epsilon\\
    \max_{a \in A \setminus \hat{A}} \hat{Q}(s',a';\hat{\theta})  \text{  otherwise}
\end{cases}$
      \State Execute action $a$ and observe reward $r$ and next state $s'$
      \State Append transition ($s,a,r,s'$) to $D$
      \State Sample random minibatch $(s_j,a_j,r_j,s'_j)$ from $D$
      \State $y_j= 
\begin{cases}
    r_j \text{ if } \mbox{final step of episode}\\
    r_j + \gamma \max_{a \in A} \hat{Q}(s',a';\hat{\theta})              & \text{otherwise}
\end{cases}$
      \State Set $err=\left( y_j-Q(s',a';\theta) \right)^2$ 
      \State Gradient descent step on $err$ with respect to $\theta$
      \State Reset $\hat{Q}=Q$ every $C$ steps
      \State $s \leftarrow$ $s'$
   \Until {$s$ is a goal state}
\Until convergence
\end{algorithmic}
\end{algorithm}

We extend the learning algorithm described in \cite{Cuayahuitl2017humanoids} by refining the action set at each time step so that agents gain access to look-ahead information and can learn to make inferences over the effects of their actions. 
In this way, our agents anticipate the effects of their decision making better than during pure naive exploration/exploitaion. An agent may for example have the winning or loosing move at some point during the game together with other available actions. But this raises the question `Why should agents learn what to do if they have the ability to infer that a game is about to be won or lost?' In the former case (win), it could omit all actions that are not winning moves---unless winning is not the objective, see line 6 in Algorithm~\ref{ourAlgorithm}. In the latter case (lose), it could avoid all actions that lead to loosing the game---see line 5 in Algorithm~\ref{ourAlgorithm}. This algorithm requires taking actions temporarily in the environment to observe the consequent rewards (with 1 look-ahead time step), and then undo such temporary actions to remain in the original environment state $s$---before looking for the worst negative actions and/or winning actions.
The main changes in contrast to the original DQN algorithm require the identification of worst actions $\hat{A}$ as well as best actions (if any) so that decision making can be made based on actions in $A$ not in $\hat{A}$, also denoted as $A \setminus \hat{A}$. Our agents thus select actions according to \begin{equation}\nonumber
\pi^{*}_\theta(s)=\arg \max_{A \setminus \hat{A}} Q^*(s,a,\theta),
\end{equation}
where both $s$ and $a$ exhibit multimodal aspects. While states $s$ include verbal and visual observations (i.e. words and game moves), actions $a$ include verbalisations and motor commands---see videos in Section~\ref{integratedSystem}. 


\section{Experimental Setting}
\label{sec:experimentalsetting}
In contrast to previous studies that require millions of video frames for training  visually-aware game-based policies \cite{mnih-dqn-2015}, we train our game move recogniser from a few hundred example images and our reinforcement learners using simulations due to the large amount of training examples required.

\subsection{Characterisation of Deep Reinforcement Learners}
\subsubsection{Multimodal States}
\begin{figure}[t!]
  \begin{center}
    \includegraphics[width=1\textwidth]{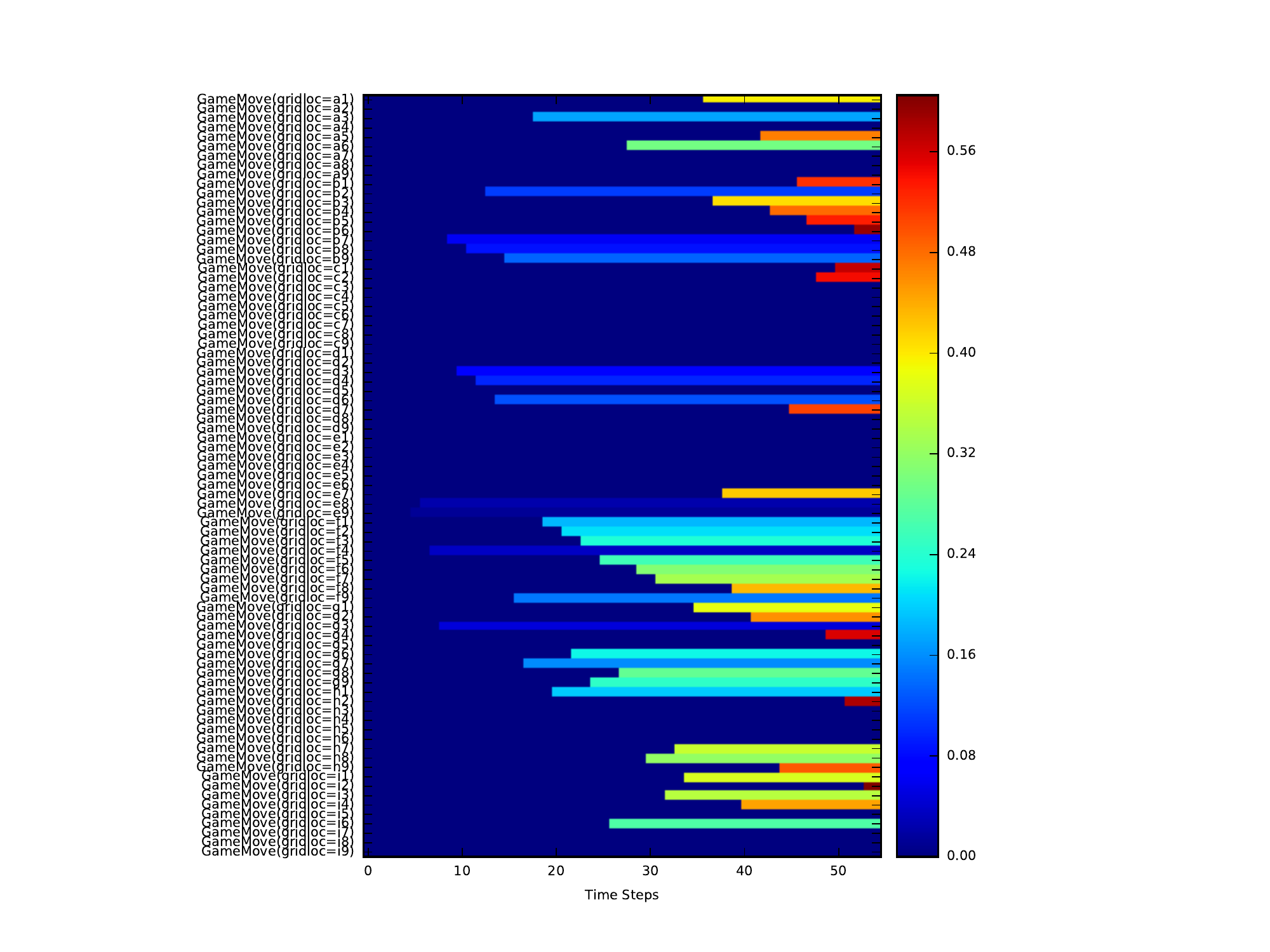}
\caption{\label{TemporalInfo}Illustration of input features describing temporal information, i.e. when game moves occur -- the higher the value the later the game move occured in a dialogue/game}
  \end{center}
\end{figure}

Our environment states include 73 and 289 features (for N\&C 3x3 and N\&C 9x9, respectively) that describe the game moves, executed commands, and words raised in the interactions. While words derived from system responses are treated as binary variables (i.e. present or absent), words derived from user responses are treated as continuous variables. We treat game moves as binary features and future work can consider treating them as continuous variables to mirror the recognition confidence. In addition, we consider a robot that cannot distinguish the order of game moves versus a robot that can distinguish the sequence of game moves. The latter is addressed by features that describe when a game move happened, where an earlier move has a lower value than a later move. We calculate such values according to $TemporalInfo=\frac{TimeStep}{|RobotGameMoves|}$, which is a value between 0 and 1. Figure~\ref{TemporalInfo} shows an example set of features and their values in a particular game, which can have different values in other games due to different sequences of game moves. Table~\ref{features} summarises the features given as inputs to our reinforcement learning agents\footnote{\textcolor{black}{Implementation wise, our states are maintained using a dictionary of key-value pairs (also known as `Hash Table')---which can be seen as a memory. From this data structure and at each time step, a vector of numerical values is generated based on a sorted list of keys for consistency purposes. In other words, every value $i$ in a different state ($s_t^i$) refers to the same key. In this way, observing a new state consists of generating a vector of numerical values from such a data structure, which is given as input to our neural network.}}.


\begin{table}[t!]
\footnotesize
\centering
\begin{tabular}{|c|l|l|l|}
\hline
\bf Num. & \bf Feature & \bf Description \\ 
\hline 
9 or 81 & $[who$=$rob \wedge what$=$draw \wedge where$=$i^*]$ & Robot game moves, one for each grid square\\
9 or 81 & $[who$=$usr \wedge what$=$draw \wedge where$=$j]$ & Human game moves, one for each grid square\\
7 & $[who$=$usr \wedge what$=$command]$ & Robot commands, e.g. gestures\\
39 & Words & Presence or absence of uttered words with \\
   &       & recognition confidence for human responses\\
9 or 81 & Temporal Information & Game moves with time-based occurrence\\
\hline
\end{tabular}
\caption{Feature sets for the standard and ultimate noughts and crosses games  corresponding to $9+9+7+39+9=73$ and $81+81+7+39+81=289$ features, respectively. While words and temporal information are continuous features $[0\dots1]$, the remaining ones are binary features $[0,1]$. \textcolor{black}{For the sake of clarity, assuming that all features are binary, the sizes of state spaces would correspond to $2^{73}$ and $2^{289}$, respectively---these sizes justify the use of a neural-based approach to scale up to such large state spaces}}
\label{features}
\end{table}

\subsubsection{Multimodal Actions}
Multimodal actions include 18 or 90 dialogue acts (for N\&C 3x3 and N\&C 9x9, respectively), where grid square $i=\{topLeft,...,bottomRight\}$ or $i=\{a1,...,i9\}$. Rather than training agents with all actions in every environment state, the actions per state were automatically restricted in two ways. First, dialogue acts are derived from the most likely actions, $Pr(a|s)>0.001$, with probabilities derived from a Naive Bayes classifier trained from example dialogues---see \cite{Cuayahuitl2017humanoids}. Second, all game moves were allowed from the subset of those not taken yet (to the robot's knowledge). Table~\ref{actions} illustrates the set(s) of outputs of reinforcement learning agents.


\begin{table}[t!]
\footnotesize
\centering
\begin{tabular}{|l|l|}
\hline
\bf Dialogue Act & \bf Multimodal Verbalisation \\ 
\hline 
Salutation(greeting) & ``Hello! [who=rob$\wedge$what=hello]"\\
Provide(name) & ``I am Pepper [who=rob$\wedge$what=please]"\\
Provide(feedback=win) & ``Yes, I won! [who=rob$\wedge$what=happy]"\\
Provide(feedback=loose) & ``No, I lost. [who=rob$\wedge$what=no]"\\
Provide(feedback=draw) & ``It's a draw. [who=rob$\wedge$what=think]"\\
GameMove(gridloc=Middle) & ``I take this one [who=rob$\wedge$what=draw$\wedge$where=middle]"\\
GameMove(gridloc=UpperMiddle) & ``I take this one [who=rob$\wedge$what=draw$\wedge$where=uppermiddle]"\\
GameMove(gridloc=LowerMiddle) & ``I take this one [who=rob$\wedge$what=draw$\wedge$where=lowermiddle]"\\
GameMove(gridloc=MiddleRight) & ``I take this one [who=rob$\wedge$what=draw$\wedge$where=middleright]"\\
GameMove(gridloc=MiddleLeft) & ``I take this one [who=rob$\wedge$what=draw$\wedge$where=middleleft]"\\
GameMove(gridloc=UpperRight) & ``I take this one [who=rob$\wedge$what=draw$\wedge$where=upperright]"\\
GameMove(gridloc=LowerRight) & ``I take this one [who=rob$\wedge$what=draw$\wedge$where=lowerright]"\\
GameMove(gridloc=UpperLeft) & ``I take this one [who=rob$\wedge$what=draw$\wedge$where=upperleft]"\\
GameMove(gridloc=LowerLeft) & ``I take this one [who=rob$\wedge$what=draw$\wedge$where=lowerleft]"\\
Request(playGame) & ``Would you like to play a game? [who=rob$\wedge$what=asr]"\\
Request(userGameMove) & ``your turn [who=rob$\wedge$what=read]"\\
Reply(playGame=yes) & ``Nice. Let me start."\\
Salutation(closing) & ``Good bye!"\\
\hline
\end{tabular}
\caption{Action set for the standard Noughts and Crosses (N\&C) game, where squared brackets denote robot commands such as gestures. A similar set is used for the ultimate N\&C game but with a larger set of moves}
\label{actions}
\end{table}

\subsubsection{State Transitions}
The features in every environment state are based on numerical vectors representing the last system and user responses, and game history. 
The language generator used template-based responses similar to those provided in Table~\ref{actions}.

\subsubsection{Rewards}
The game-based rewards are as follows:
\begin{equation}\nonumber
r(s,a,s')= 
\begin{cases}
    +5 \text{ for } \mbox{winning or about to win the game}\\
    +1 \text{ for } \mbox{a draw or about to draw in the game}\\
    -5 \text{ for } \mbox{a repeated (already taken) action}\\
    -5 \text{ for } \mbox{loosing or about to loose the game}\\
    0 \text{ otherwise.}
\end{cases}
\end{equation}
 
\subsubsection{Model Architectures}
The neural networks consist of fully-connected multilayer neural nets with 5 layers organised as follows: 62 or 207 nodes in the input layer, 100 nodes in each hidden layer, and 18 or 90 nodes in the output layer. The hidden layers use ReLU (Rectified Linear Units) activation functions to normalise their weights. The same learning parameters are used for both games including  experience replay size=10000, burning steps=1000, discount factor=0.7, minimum epsilon=0.005, batch size=2, learning steps=200K, and maxium number of actions per dialogue=100. 


\subsubsection{Simulated Interactions}
In our simulated dialogues (one per game) and for practical purposes, the behaviour of the simulated opponent is driven by semi-random user behaviour, i.e. from random but legal game moves. While system actions are chosen by the learnt policies, system responses are sampled from templates (verbalisations seen in demonstration dialogues). In addition, while non-game user actions are sampled from observed interactions in the demonstration dialogues, game user actions are generated randomly from available legal actions in order to explore all possible game strategies.

\subsection{Integrated System}
\label{integratedSystem}
The humanoid robot `Pepper'\footnote{\url{http://www.aldebaran.com/en/a-robots/who-is-pepper}} was equipped with the components below running concurrently, via multi-threading. This robot system uses the Naoqi API version 2.5, and has been fully implemented and tested. Example interactions can be seen in the following videos: \url{https://www.youtube.com/watch?v=8MqBdkfNl4c} and \url{https://www.youtube.com/watch?v=377tVIvd67I}. While the former uses handwriting, the latter does not use it due to higher complexity (future work)---instead, the human opponent does the handwriting on behalf of the robot.

\subsubsection{Interaction Manager}
The interaction manager, based on the publicly available tools {\it SimpleDS}\footnote{\url{https://github.com/cuayahuitl/SimpleDS}} \cite{Cuayahuitl2017} and ConvNetJS\footnote{\url{http://cs.stanford.edu/people/karpathy/convnetjs}}, can be seen as the robot's brain due to orchestrating modalities by continuously receiving visual and verbal perceptions from the environment, and deciding what to do next and when based on the learning agents described above. Most actions are multimodal; for example, action $GameMove(gridloc=a1)$ can be unfolded as ``I take this one $[who$=$rob \wedge what$=$draw \wedge where$=$a1]$'', where the square brackets denote a physical action (drawing a circle or cross at the given location). 

\subsubsection{Speech Recognition}
This component activates the Nuance Speech Recogniser once the robot finishes speaking. 
Although the targeted games are mostly based on non-verbal interaction, speech recognition results (words with confidence scores) are used as features in the state space of the deep reinforcement learning agents.

\subsubsection{Game Move Recognition}
This component receives video frames as input in order to output interpreted game moves as described in Section~\ref{Learning2Perceive}. It gets active as soon as previous verbal and non-verbal commands are executed, and inactive as soon as a game move is recognised. In other words, our robot uses an automatic turn-taking mechanism based on recognised game moves. These vision-based perceptions are used as features in the state space of the deep reinforcement learners. 


\subsubsection{Speech Synthesis}
The verbalisations in English, translations of high-level actions from the interaction manager, used a template-based approach and the built-in Acapela speech synthesizer. They were synchronised with arm movements, where the next verbalisation waited until the previous verbalisation and arm movements completed their execution.

\subsubsection{Arm Movements and Wheel-Based Locomotion}
This component receives commands from the interaction manager for carrying out gestures. We used both built-in arm movements for non-game moves and pre-recorded arm movements from human demonstrations to indicate game moves. In addition and due to the robot's short arms, it used its omnidirectional wheels to move from an initial location (right in front of the game grid) to the left/right/front/back in order to reach a targeted grid cell in the game---with return to the initial location. While our robot used locomotion for the standard N\&C game, it only indicated verbally its game moves in the case of the ultimate N\&C game. The latter  was due to higher complexity of motor commands and interaction efficiency without locomotion. This game setting required the human player to do the drawings on behalf of the robot. 



\section{Automatic Evaluation}
\label{intrinsiceval}


\begin{figure}[t!]
     \begin{center}
        \subfigure[Training examples without noise]{%
            \includegraphics[width=0.85\textwidth]{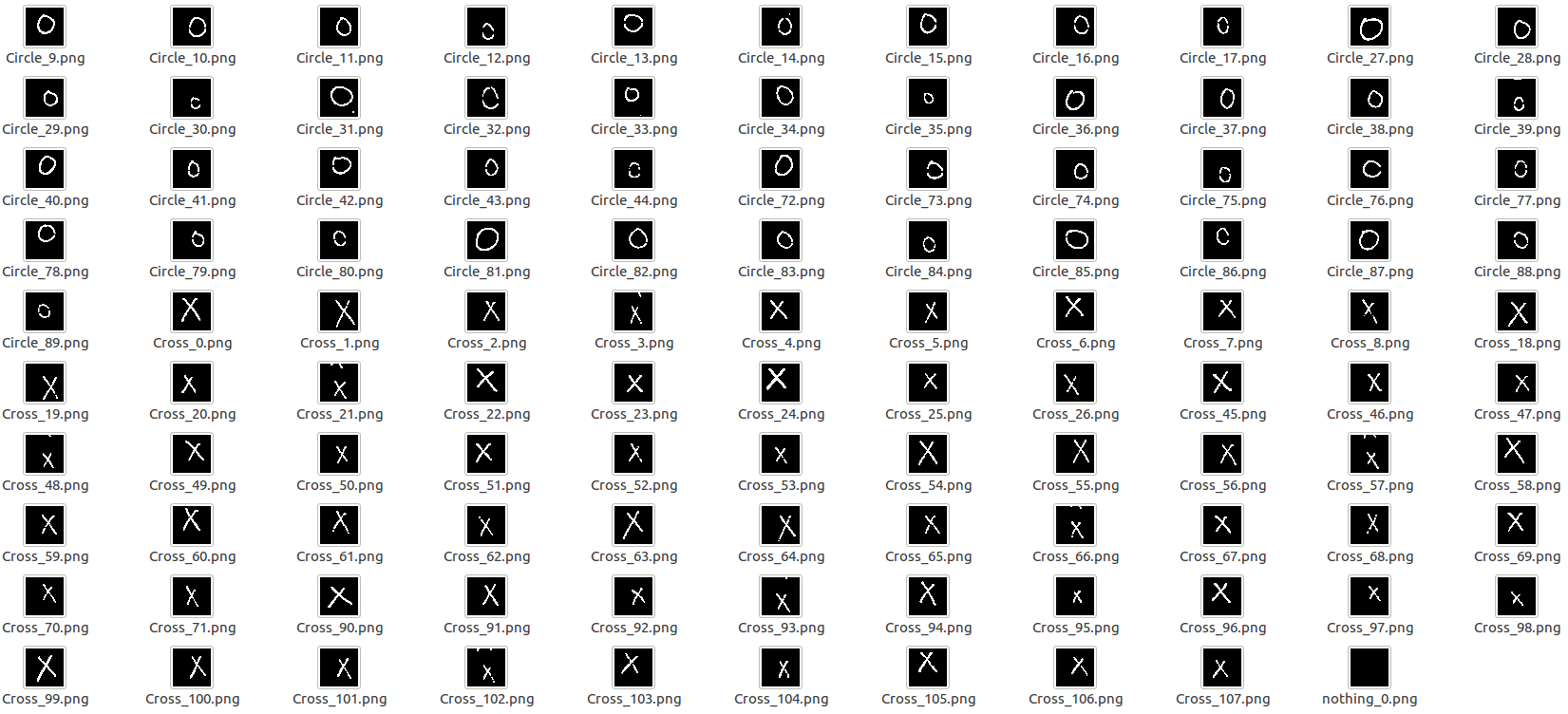}
        }
        \hspace{-0.3cm}
        \subfigure[Representative examples with noise]{
            \includegraphics[width=0.85\textwidth]{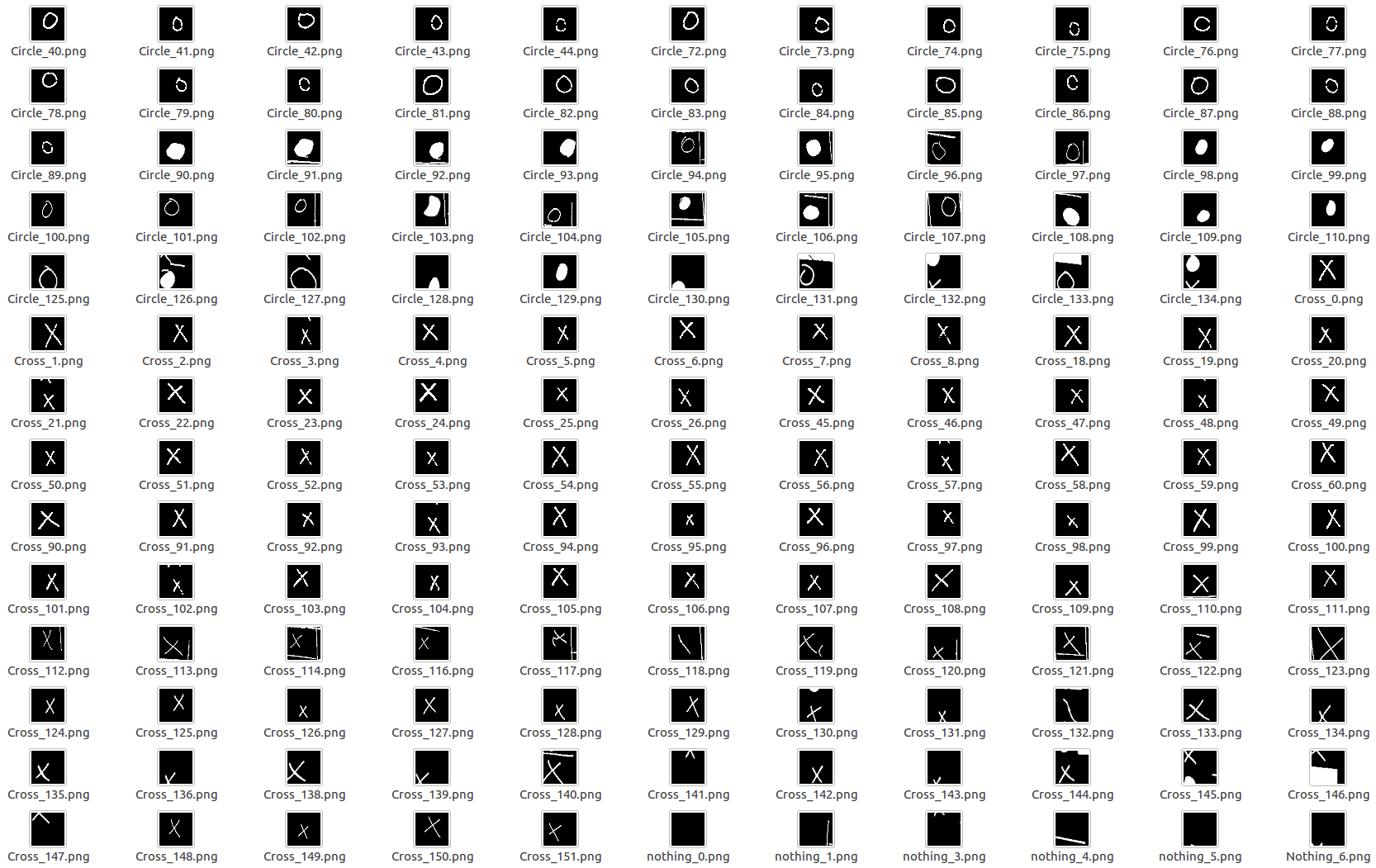}
        }
    \end{center}
    \caption{Example images for training the game move recogniser}
   \label{images}
\end{figure}

\subsection{Deep Supervised Learner for Character Recognition}
This classifier labels grayscale images into three classes (nought, cross, nothing) as described in Section~\ref{Learning2Perceive}. The  classifier used two sets of images: one set without noise (109 images as shown see Figure~\ref{images}(a)), and the other set with noise (201 images from human writings with partially included grid lines as shown in Figure~\ref{images}(b)). First, the classification accuracy in the data set without noise was $99.9\%$ according to a leave-one-out cross validation. 
Second, 
the classifier trained without the noisy data set obtained $74\%$ of classification accuracy when tested in the noisy data set. 
Third, the classifier trained with the noisy data set obtained $98.4\%$ of classification accuracy when tested in the non-noisy data set---see more details in Figure~\ref{confusionMatrix}. This is an indication of accurate classification of human handwriting for the targeted game. At the same time though these results suggest that a vision-based classifier should be retrained in case substantially different images are observed.

\begin{table}[t!]\small
\label{confusionMatrix}
\small
\centering
\begin{tabular}[c]{|c|c|c|c| }
 \hline
  & Nought & Cross & Nothing\\
 \hline
Nought & 95.1\% & 0 & 0\\
Cross & 4.9\% & 100\% & 0\\
Nothing & 0 & 0 & 100\%\\
 \hline
\end{tabular}
\caption{Confusion matrix of test results in character recognition} 
\end{table}

\begin{figure*}[t!]
     \begin{center}
        \subfigure[DQN-Original (baseline 1)]{%
            \includegraphics[width=0.45\textwidth]{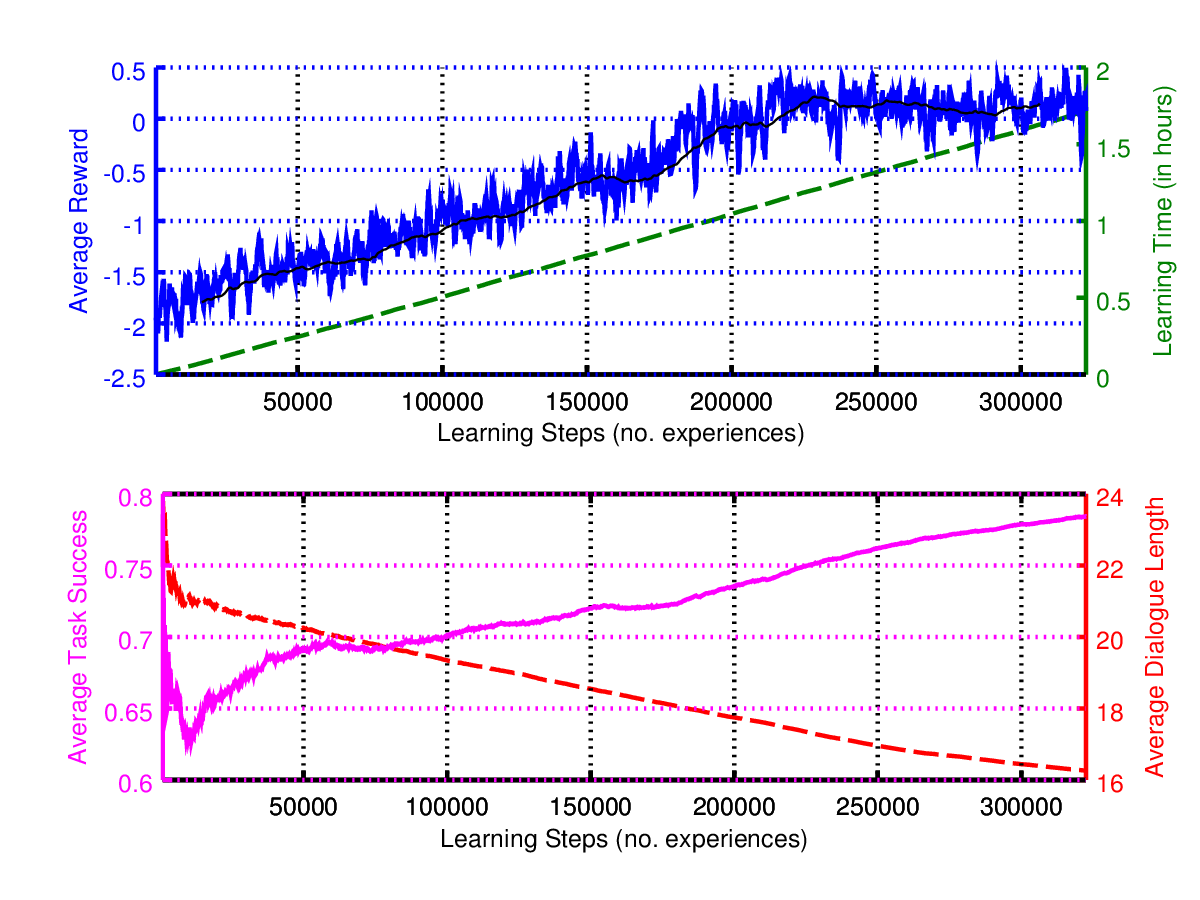}
        }
        \subfigure[DQN-Variant (baseline 2)]{%
            \includegraphics[width=0.45\textwidth]{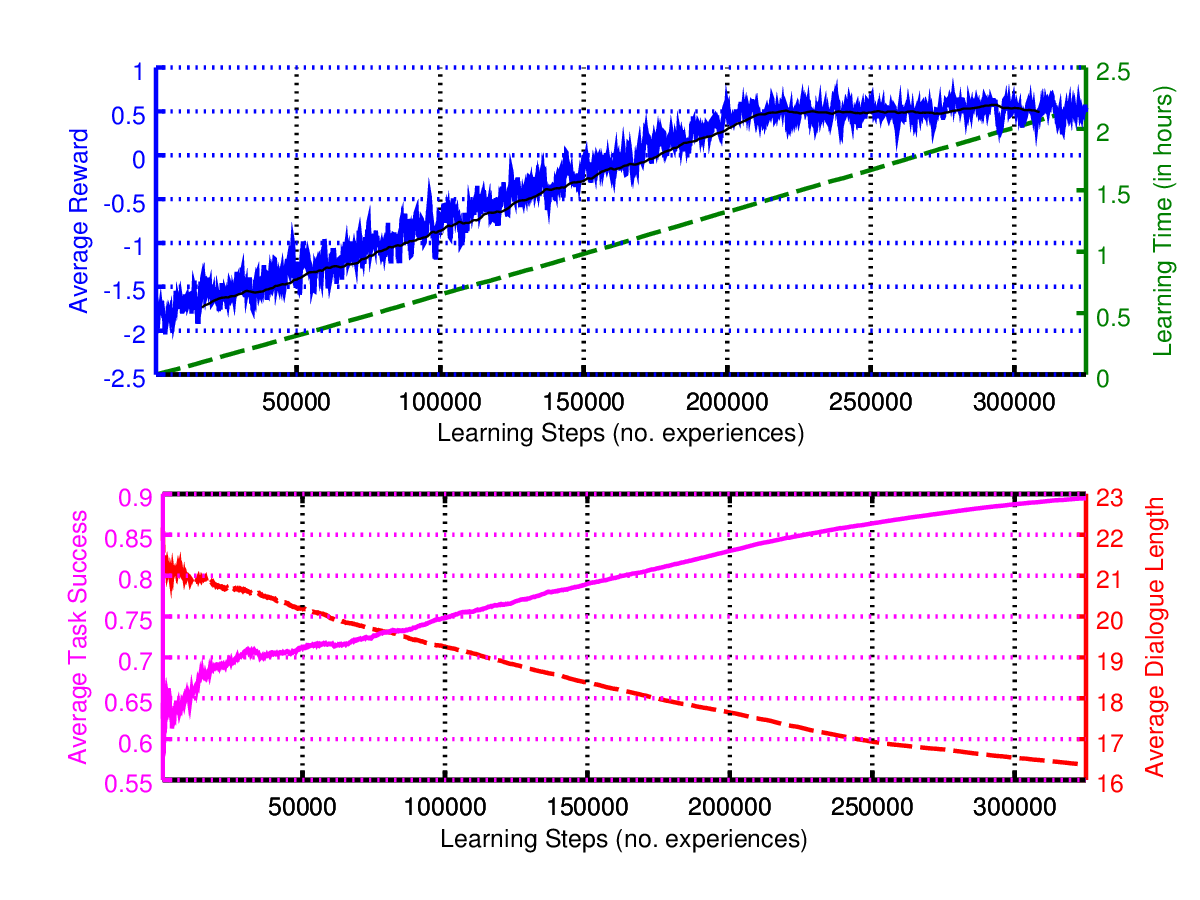}
        }
        \subfigure[Proposed without Temporal Info.]{%
            \includegraphics[width=0.45\textwidth]{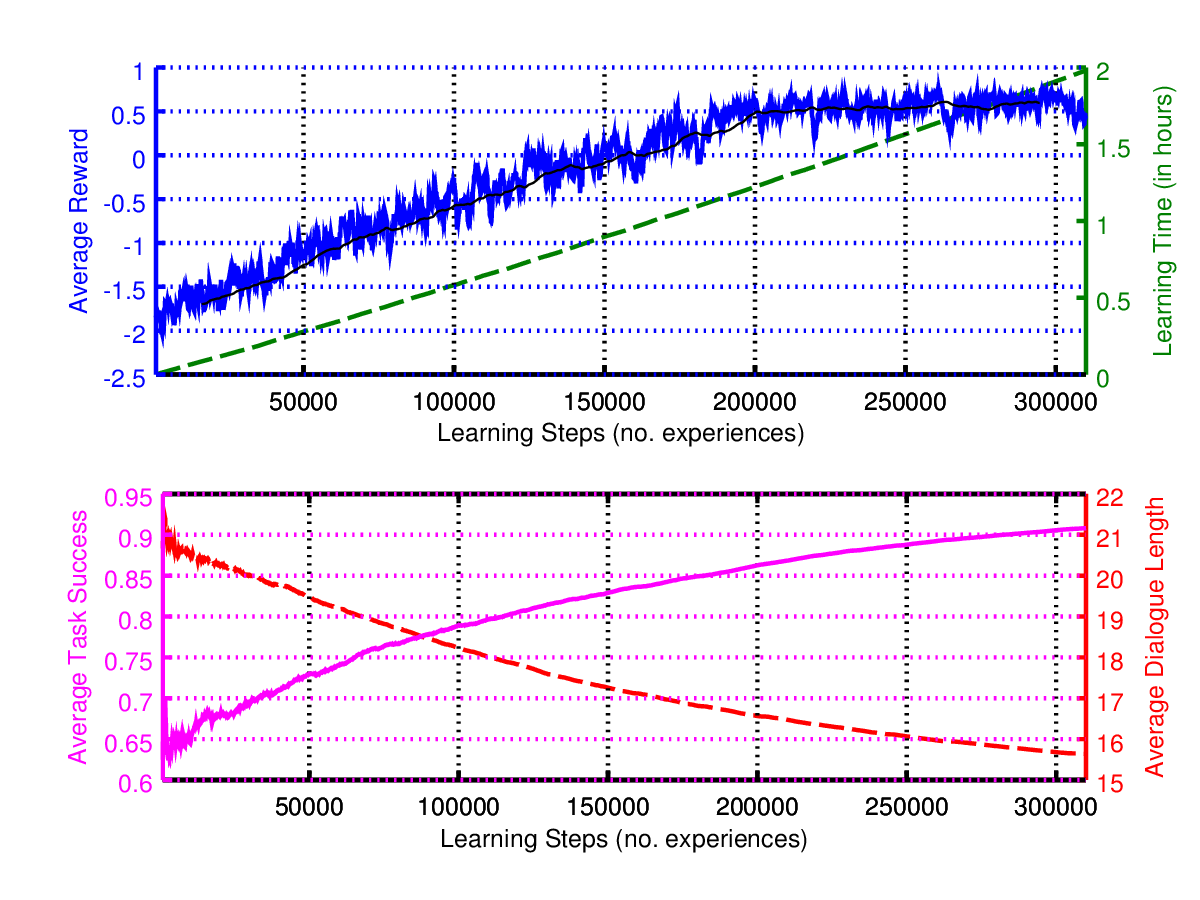}
        }
        \subfigure[Proposed with Temporal Info.]{%
            \includegraphics[width=0.45\textwidth]{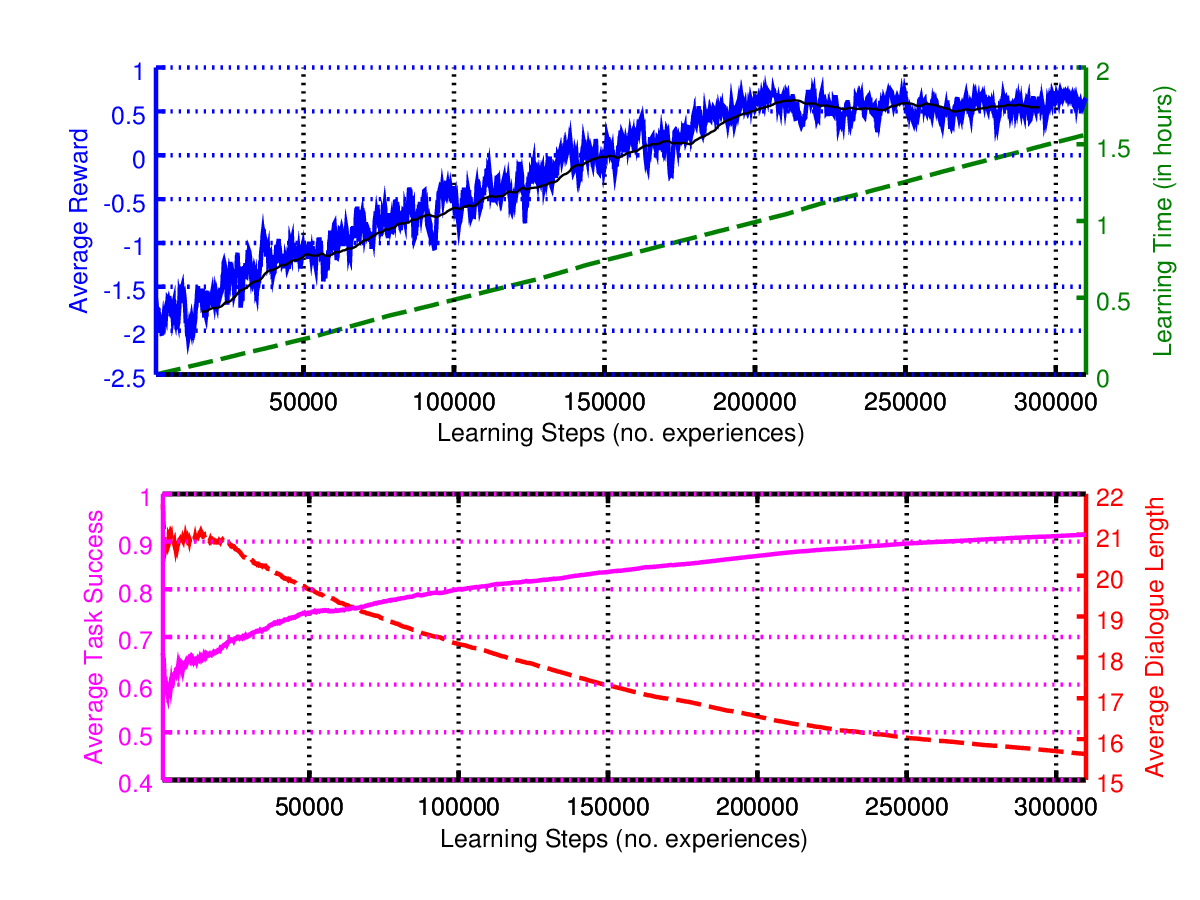}
        }        
    \end{center}
   \caption{Learning curves of DQN-based agents for playing Standard Noughts and Crosses}
   \label{learningCurves}
\end{figure*}

\begin{figure*}[t!]
     \begin{center}
        \subfigure[DQN-Original (baseline 1)]{%
            \includegraphics[width=0.45\textwidth]{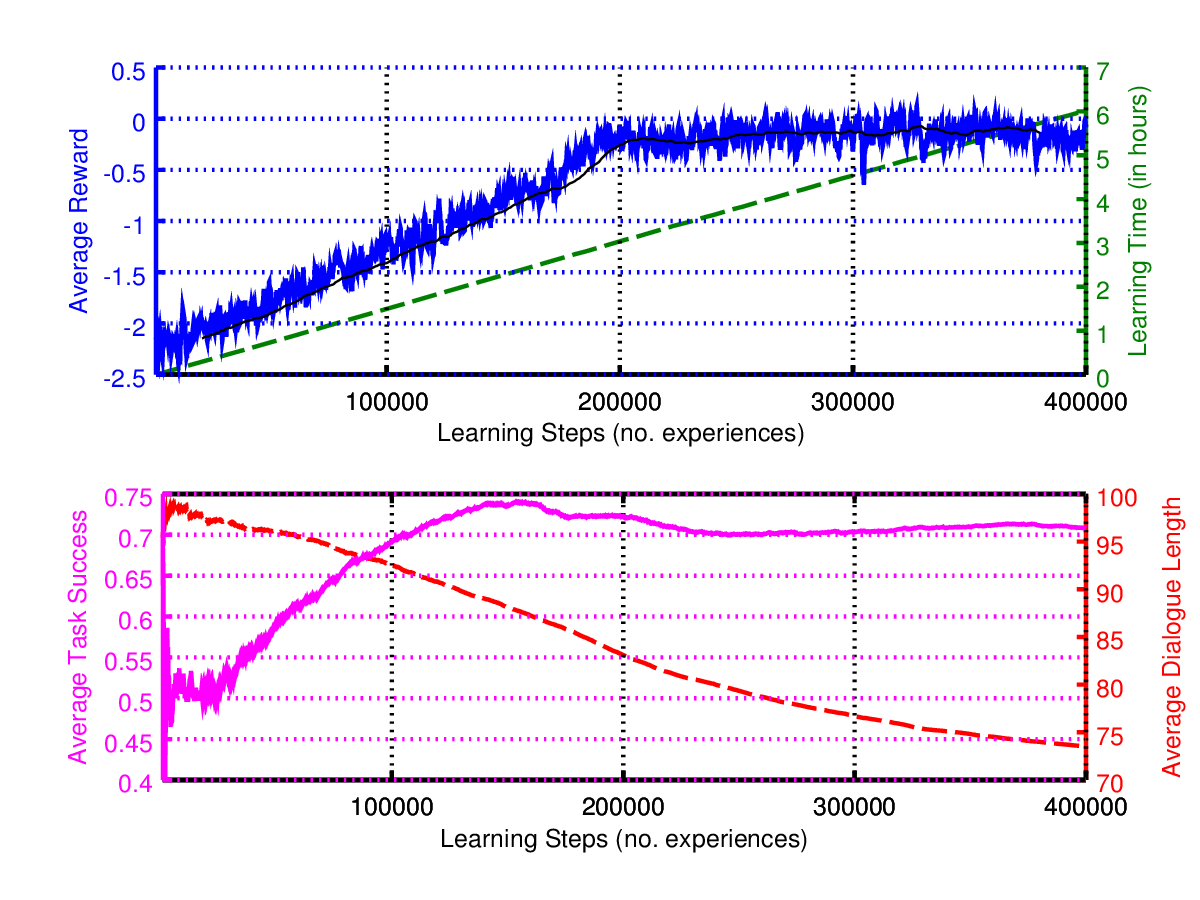}
        }
        \subfigure[DQN-Variant (baseline 2)]{%
            \includegraphics[width=0.45\textwidth]{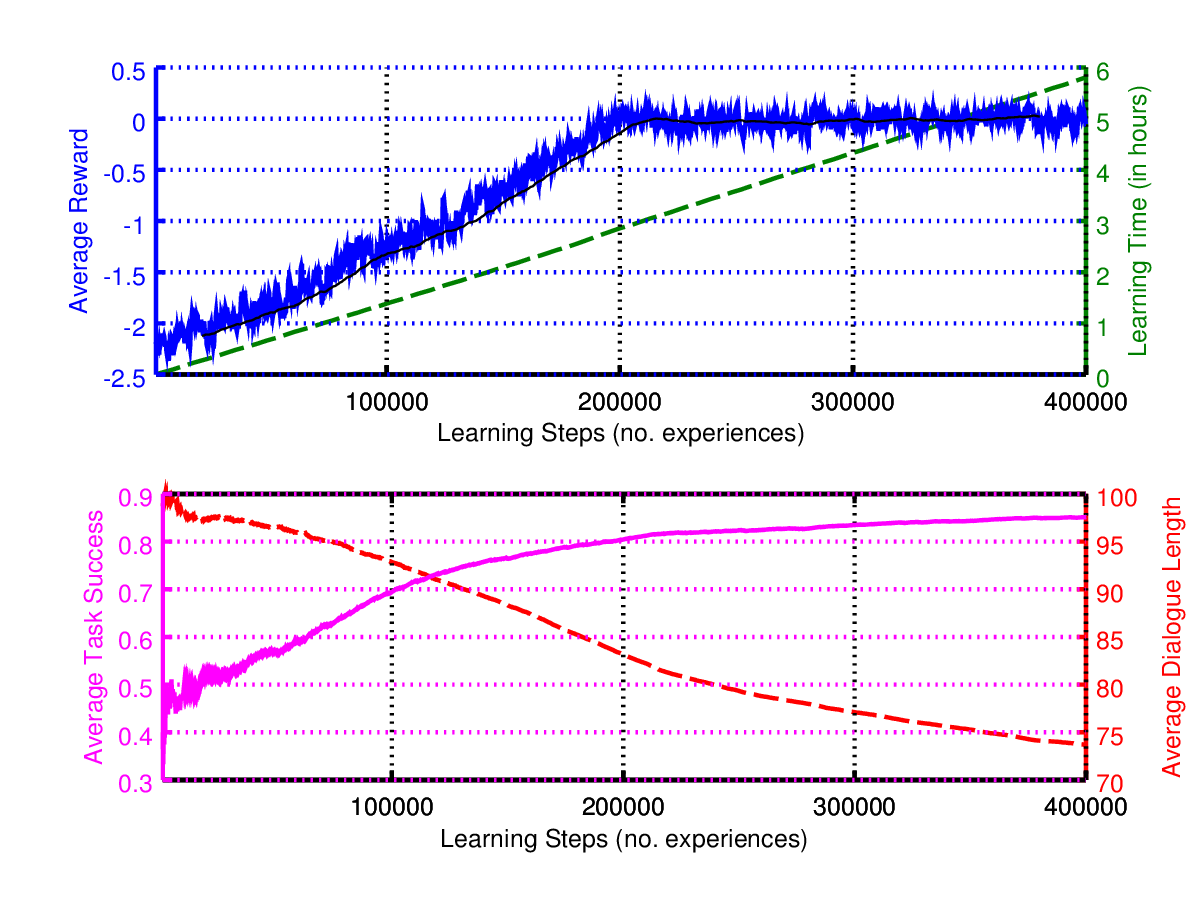}
        }
        \subfigure[Proposed without Temporal Info.]{%
            \includegraphics[width=0.45\textwidth]{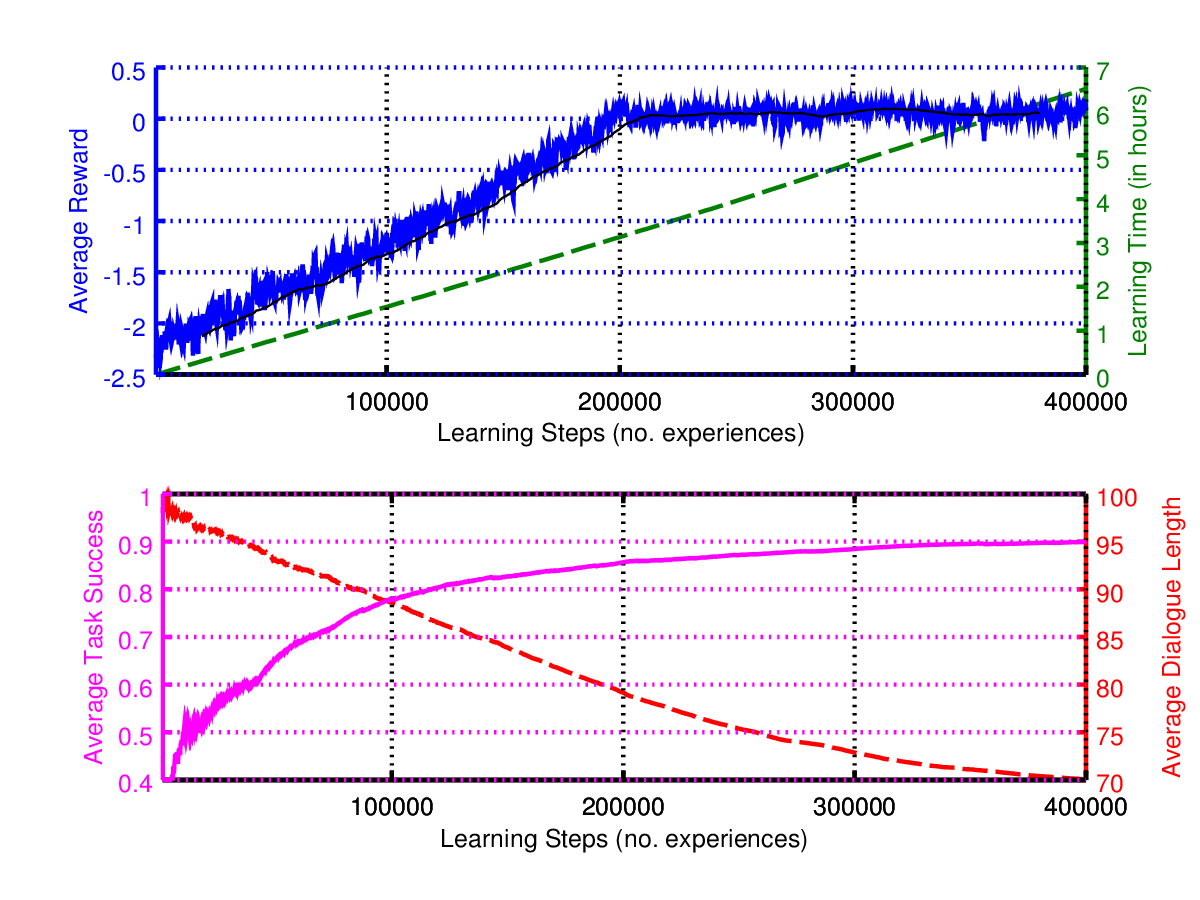}
        }
        \subfigure[Proposed with Temporal Info.]{%
            \includegraphics[width=0.45\textwidth]{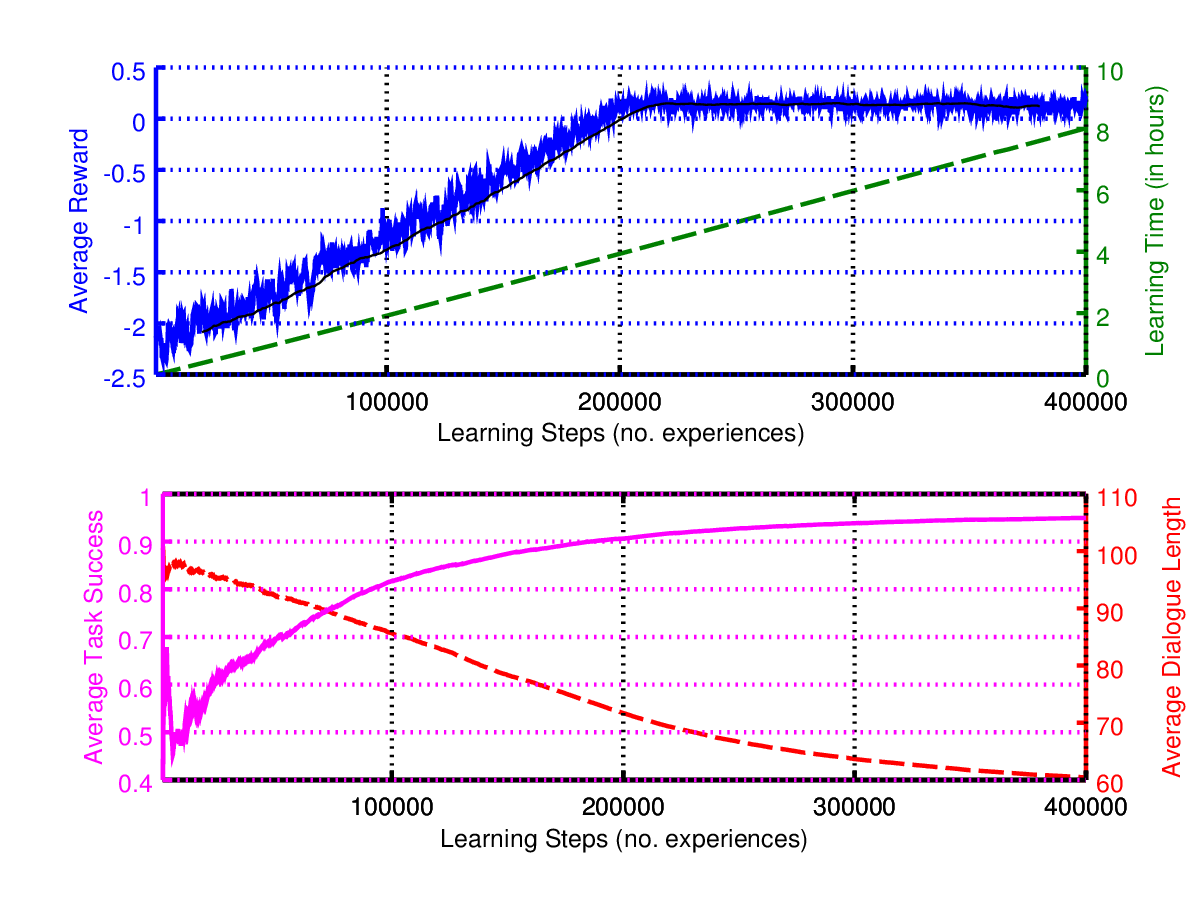}
        }
    \end{center}
   \caption{Learning curves of DQN-based agents for playing Ultimate Noughts and Crosses}
   \label{learningCurves2}
\end{figure*}

\subsection{Deep Reinforcement Learners for Game Playing}
We compare our proposed algorithm described in Section~\ref{Learning2Interact} against two baselines \cite{mnih-dqn-2015,Cuayahuitl2017humanoids} in the domain of Noughts and Crosses (N\&C) with two variants. 
Figures~\ref{learningCurves}~and~\ref{learningCurves2} show learning curves for the baseline agents (see top plots (a) and (b)), and agents using the proposed algorithm (see bottom plots (c) and (d)). All agents report results over 400K learning steps or 20K games---whatever occurred first. We use four metrics to measure system performance: average reward, learning time\footnote{Ran on Intel Core i5-3210M CPU$@$2.5GHzx4; 8GB RAM$@$2.4GHz.}, average task success $[0...1]$ (win/draw rate), and avg. dialogue length (avg. number of actions per game). Results can be seen as the higher the better in avg. reward and avg. task success, and the lower the better in training time and dialogue length.

Our results show that our proposed algorithm can train more successful agents than previous work. This is evidenced by higher task success in plots (c) and (d) vs. (a) and (b), and lower dialogue length in plots (c) and (d) vs. (a) and (b).


We tested the performance of the learnt policies over 3000 games for each of the three agents per game and per architecture (100 vs. 150 nodes per hidden layer), obtaining the results shown in Table~\ref{testResults}. It can be noted that indeed the proposed learning algorithm performs better than the baseline algorithms, across games and model architectures. These results also suggest that there is room for hyperparameter optimisation in future work. 
Nonetheless, these results suggest that our proposed algorithm can be used for training agents with competitive behaviour in social games. 


\begin{table}[t!]\small
\small
\centering
\begin{tabular}[c]{|c|l|c|c|c| }
 \hline
Game$^{ModelArch.}$ & Learning & Average & Task  & Dialogue\\
     & Algorithm & Reward & Success & Length \\
 \hline
 \hline
\multirow{ 3}{*}{Standard N\&C$^1$} & DQN-Original (Baseline 1) \cite{mnih-dqn-2015} & 0.0658 & 0.8258 & 13.93 \\
 & DQN-Variant (Baseline 2) \cite{Cuayahuitl2017humanoids} & 0.5530 & 0.9720 & 13.99 \\
 & Proposed without Temporal Info. & 0.4710 & 0.9868 & 15.31 \\
 & Proposed with Temporal Info. & 0.6300 & {\bf 0.9980} & 14.06 \\
\hline 
\multirow{ 3}{*}{Ultimate N\&C$^1$} & DQN-Original (Baseline 1) \cite{mnih-dqn-2015} & -0.6900 & 0.7873 & 63.54 \\
 & DQN-Variant (Baseline 2) \cite{Cuayahuitl2017humanoids} & 0.0177 & 0.9074 & 64.82 \\
 & Proposed without Temporal Info. & 0.0693 & 0.9377 & 63.62 \\
 & Proposed with Temporal Info. & 0.1440 & {\bf 0.9753} & 52.03 \\
 \hline
 \multirow{ 3}{*}{Ultimate N\&C$^2$} & DQN-Original (Baseline 1) \cite{mnih-dqn-2015} & -0.1120 & 0.7290 & 65.49 \\
 & DQN-Variant (Baseline 2) \cite{Cuayahuitl2017humanoids} & -0.0310 & 0.8663 & 70.57 \\
 & Proposed without Temporal Info. & 0.0997 & 0.9473 & 59.69 \\
 & Proposed with Temporal Info. & 0.1640 & {\bf 0.9846} & 52.09 \\
 \hline
\end{tabular}
\caption{Test results averaged over 3000 games (N\&C=Noughts and Crosses) using the baseline and proposed algorithms. $^1$ and $^2$ used 100 and 150 nodes per hidden layer, respectively} 
\label{testResults}
\end{table}

\section{Human Evaluation}
\label{extrinsiceval}
\begin{figure}[t!]
     \begin{center}
             \subfigure[Training and test environment]{%
            \includegraphics[width=0.45\textwidth]{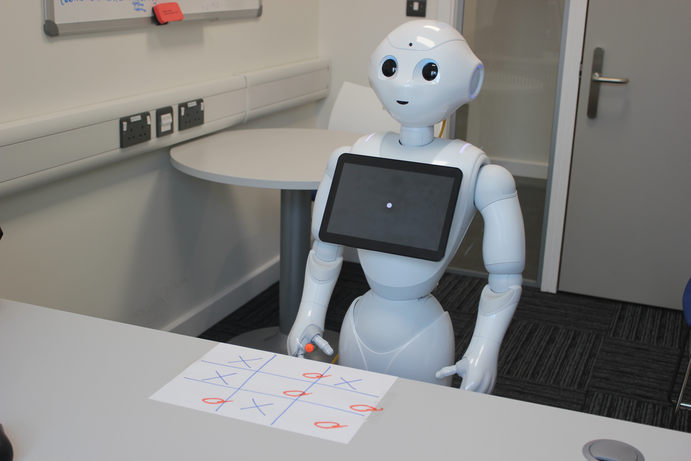}
        }
        \hspace{-0.15cm}
        \subfigure[Deployment environment.]{%
            \includegraphics[width=0.534\textwidth]{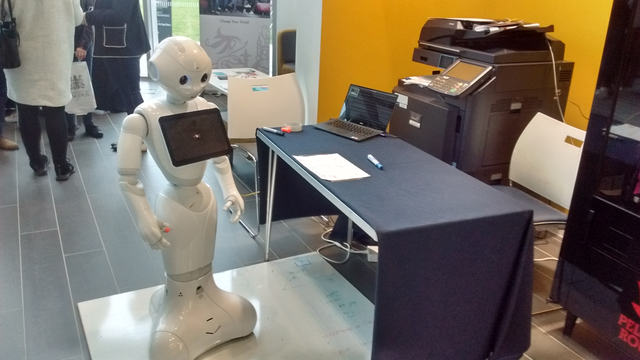}
        }
    \end{center}
   \caption{Robot's training, test, and deployment environments}
   \label{environments}
\end{figure}

We trained and tested our robot system in an office environment, and deployed it in a partially-known environment (atrium of a University building)---see Figure~\ref{environments}. 
This evaluation ran for four non-consecutive days where the robot  played against 29, 64, 27 and 10 human opponents, respectively. The first three days involved only the Standard N\&C game, and the latter involved both games (Standard N\&C and Ultimate N\&C). These human-robot games included primary and secondary school children, prospective university students, and parents---they were visitors to the building rather than traditional experiment participants (no questionnaires involved). While our best interaction policies (but without temporal information) were used over all games across days, the game move recogniser evolved due to improvements after each deployment day. The improvements consisted in better game grid detection and hand detection, which reduced misrecognised game moves as follows: 31\% on day one, 25\% on day two, 22\%, and 10\% on day four. In games without misrecognitions the robot ended up winning or in a draw. Algorithm~\ref{GMR} describes our game move recogniser after such improvements, which has been used in both games (Standard and Ultimate Noughts and Crosses) and can be applied to other social games beyond those in this article. Algorithm~\ref{ndqn} has also been used in both games and can be applied to other social games and domains beyond those in this article. The reason for this is because this algorithm is general enough, not only applicable to games, as long as there is a notion of success and failure (in our case win or loose) at the end of a dialogue, where the same reward function can be applied or extended. 

Although the partially-known conditions exhibited in the deployment environment were challenging for our interactive multimodal robot (e.g. different light conditions, multiple visual backgrounds, two human opponents instead of one, among others), its human opponents continuously expressed to be impressed -- presumably due to the fact that the robot was speaking, moving, writing, listening, and seeing---all of these autonomously, with near real-time execution, and in a coordinated way. 

\section{Conclusion and Future Work}
\textcolor{black}{This article presents a deep learning-based approach for efficiently training deployable robots that can carry out joint multimodal activities together with humans that involve speaking, listening, gesturing, and learning. Although the proposed approach assumes no initial training data, it is bootstrapped with relatively small datasets, and it learns interactive behaviour from trial and error using simulations. This approach is demonstrated using the game of Noughts and Crosses (N\&C) with two variants. Even when these two variants exhibit different degrees of complexity, the data requirements remained equivalent. In other words, the more complex task (Ultimate N\&C) did not require more data than the simpler task (Standard N\&C). Given the generality of the approach and proposed algorithms, they can also be applied to other tasks beyond N\&C.} 
An automatic evaluation shows that our deep supervised and reinforcement learners achieve high performance in both game move recognition and task success. In the latter, anticipating the effects of the decision making and temporal information proved essential for improved performance. Our experimental results with 130 human participants showed that when the vision-based perception works as expected, successful human-robot interactions can be obtained from the induced skills using the proposed data-efficient approach. 

Example avenues for future related works are as follows.
\begin{itemize}
\item \textcolor{black}{Extending the robot's language skills for larger-scale language interpretation \cite{Deng2018} and language generation \cite{Dethlefs17} coupled with visual perception, multimodal interaction and motor commands remains to be investigated. In addition, the conversational behaviour of the robot can be framed within a chatbot approach \cite{CuayahuitlEtAl2018slt} in order to deal with the out-of-domain responses pointed out by \cite{Bohus_2014}.}
\item A comparison of the temporal information approach used in this article versus an approach based on recurrent neural networks (as in \cite{QureshiNYI17a}) remains to be investigated, identifying pros and cons of each approach. 
\item Another interesting extension to this work is {\it online training} \cite{Settles2012,CuayahuitlDethlefs2012}, which can be investigated for improving the performance of both supervised and reinforcement learning by for example retraining them after each game. For this, it should be taken into account that the reinforcement learner requires longer training times than its supervised counterpart, where faster training algorithms can be investigated by combining ideas from \cite{NairSBAFMPSBPLM15,MnihBMGLHSK16,SchaulQAS15,HeLSP16,KulkarniNST16,CuayahuitlEtAl2017ijcnn} and \cite{CuayahuitlYu2017}.
\item \textcolor{black}{Another interesting extension is {\it learning to write} as in \cite{GregorDGRW15,BullockGM93}. While handwriting was relatively straightforward for the standard Nought and Crosses game, handwriting for the ultimate Noughts and Crosses game was a challenge due to more fine grained motions of smaller characters in smaller grid squares on the game board. This extension requires visuomotor learning in order to achieve human-like  handwriting.}
\item \textcolor{black}{Our policy learning algorithm with 1 step look-ahead information could be explored with multiple time steps, leading to combinations of deep reinforcement learning policies with MinMax \cite{KallesKanellopoulos2008} or Monte Carlo tree search methods \cite{silver2016alphago}. But there is a trade-off because the more look-ahead information is used, the more computational expense is involved. The training times of our proposed algorithm vs. the original DQN method are equivalent, and the quality of policies are substantially different---ours much more competitive. In addition, our policy learning algorithm can be applied for optimising robot behaviour that goes beyond always winning. Instead, it could be used to train multimodal robots that keep players as happy as possible. This would require adding further multimodal actions (for varied behaviour rather than repetitive or monotonous behaviour across games) and also keeping track of emotion-based signals, which can be taken into the reward function for policy retraining.}
\item The proposed algorithms can be applied to other social games (possibly more complex) and also other domains. A robot at home for example should not only be expected to be able to play games, but also to carry out other tasks---potentially trained with the same algorithms across tasks.
\item More real-world evaluations are needed across the field to truly and thoroughly assess the performance of human-robot interactions in the wild, out of the lab \cite{HC2015aisb,Jung2018,Cuayahuitl2015csl}. The robot system described in this article would not have been possible without the participation of unseen humans playing against the robot.
\end{itemize}

\section*{Acknowledgement}
The robot used in this paper was donated by the Engineering \& Physical Sciences Research Council (EPSRC), U.K. This work was carried out under ethical approval by the University of Lincoln with reference UID CoSREC396.


\bibliography{hc-neurocomp2018}

\end{document}